%% file: main.tex
\documentclass[10pt,twocolumn,letterpaper]{article}
\usepackage{balance}
\usepackage{flushend}
\usepackage{amssymb}
\usepackage{amsmath}
\usepackage[pagenumbers]{cvpr}
\usepackage[table]{xcolor}
\usepackage{pifont}
\usepackage{color}
\usepackage[accsupp]{axessibility}
\usepackage{xcolor}

\definecolor{best}{RGB}{241,65,23}
\definecolor{second}{RGB}{153,51,0}
\definecolor{third}{RGB}{255,255,0}
\definecolor{darkRed}{RGB}{241,65,23}
\usepackage{fontawesome}
\input{preamble}

\definecolor{cvprblue}{rgb}{0.21,0.49,0.74}
\usepackage[pagebackref,breaklinks,colorlinks,citecolor=cvprblue,urlcolor=cvprblue]{hyperref}
\vspace{-15pt}\title{\Large Quadratic Gaussian Splatting: High Quality Surface Reconstruction with Second-order Geometric Primitives\vspace{-5pt}}

\author{
    {Ziyu Zhang$^{1,2,\dagger}$, Binbin Huang$^{3,\dagger}$, Hanqing Jiang$^4$, Liyang Zhou$^4$, Xiaojun Xiang$^4$, Shuhan Shen$^{1,2,*}$}\vspace{8pt}\\
    $^1$CASIA \quad $^2$UCAS \quad $^3$The University of Hong Kong \quad $^4$SenseTime Research\vspace{8pt}\\
    \large{\url{https://quadraticgs.github.io/QGS}}
}
\begin{document}
\maketitle

\input{sec/0_abstract}
\begin{figure}
    \centering
    \begin{minipage}{0.24\linewidth}
        \includegraphics[width=1\linewidth]{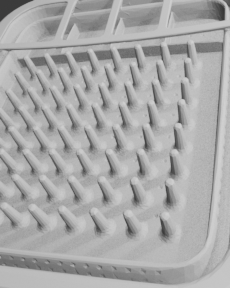}
    \vspace{-18pt}
    \caption*{GT}
    \end{minipage}
    \begin{minipage}{0.24\linewidth}
        \includegraphics[width=1\linewidth]{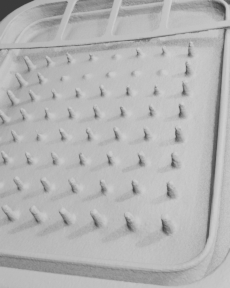}
    \vspace{-18pt}
    \caption*{2DGS}
    \end{minipage}
    \begin{minipage}{0.24\linewidth}
        \includegraphics[width=1\linewidth]{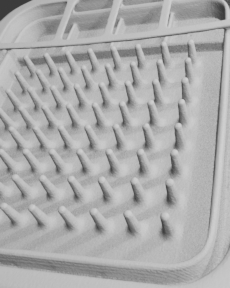}
    \vspace{-18pt}
    \caption*{PGSR}
    \end{minipage}
    \begin{minipage}{0.24\linewidth}
        \includegraphics[width=1\linewidth]{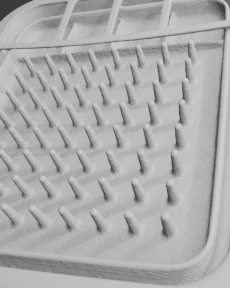}
    \vspace{-18pt}
    \caption*{QGS (8K)}
    \end{minipage}
    \centering
    \includegraphics[width=1\linewidth]{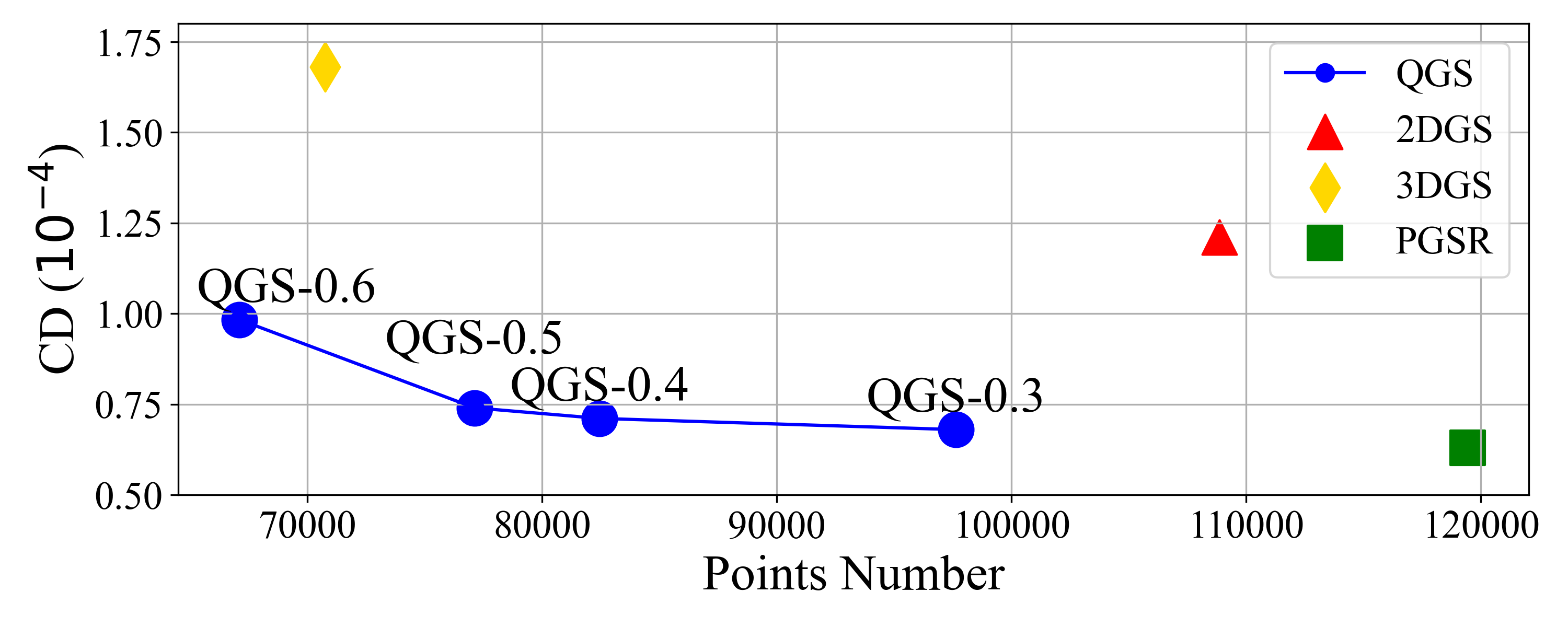}
    \vspace{-20pt}
    \caption{Our QGS model employs quadric surfaces to recover high-precision geometry from multi-view RGB images, reducing point count while enhancing accuracy. To isolate geometry from texture, we evaluate QGS on a weakly textured, geometry-rich subset of the Google Scanned Object dataset~\cite{downs2022google}. The results suggest that QGS, under different gradient threshold settings (QGS-x), consistently outperforms 2DGS~\cite{huang20242d} and achieves comparable quality to PGSR~\cite{chen2024pgsr} with fewer points.}
    \label{fig:teaser}
    \vspace{-10pt}
\end{figure}

\input{sec/1_intro}

\input{sec/2_relatedwork}

\input{sec/3_method}

\input{sec/4_experiments}
\input{sec/5_conclusion}
\clearpage
{
    
    \small
    \bibliographystyle{ieeenat_fullname}
    \bibliography{main}
    
}

\input{sec/X_suppl}

    
    

\end{document}

%% file: preamble.tex
\usepackage{longtable}
\usepackage{graphicx}
\usepackage{tikz}
\usepackage{graphicx}
\usepackage{multirow}
\usepackage{subcaption}
\usepackage{lipsum}
\usepackage{cuted}
\usepackage{indentfirst}


%% file: sec/0_abstract.tex
\begin{abstract}
We propose Quadratic Gaussian Splatting (QGS), a novel representation that replaces static primitives with deformable quadric surfaces (e.g., ellipse, paraboloids) to capture intricate geometry. Unlike prior works that rely on Euclidean distance for primitive density modeling—a metric misaligned with surface geometry under deformation—QGS introduces geodesic distance-based density distributions. This innovation ensures that density weights adapt intrinsically to the primitive curvature, preserving consistency during shape changes (e.g., from planar disks to curved paraboloids). By solving geodesic distances in closed form on quadric surfaces, QGS enables surface-aware splatting, where a single primitive can represent complex curvature that previously required dozens of planar surfels, potentially reducing memory usage while maintaining efficient rendering via fast ray-quadric intersection. Experiments on DTU, Tanks and Temples, and MipNeRF360 datasets demonstrate state-of-the-art surface reconstruction, with QGS reducing geometric error (chamfer distance) by 33\% over 2DGS and 27\% over GOF on the DTU dataset. Crucially, QGS retains competitive appearance quality, bridging the gap between geometric precision and visual fidelity for applications like robotics and immersive reality. 
\vspace{-10pt}
\end{abstract}

%% file: sec/1_intro.tex
\section{Introduction}
\label{sec:introduction}

The pursuit of photorealistic view synthesis and geometry reconstruction is a central focus in academia and industry, with applications spanning virtual reality, film-making, and autonomous driving. Recent point-based methods, such as 3D Gaussians~\cite{kerbl3Dgaussians, yu2024gaussian}, have proven highly effective, delivering view synthesis quality comparable to Neural Radiance Fields (NeRF)-based approaches while preserving superior geometry~\cite{huang20242d, DSS} and offering significantly faster performance. Since then, these methods have been extended to dynamic reconstruction~\cite{huang2024endo, feng2024gaussian, lin2024gaussian}, scene editing~\cite{chen2024gaussianeditor, waczynska2024games, tobiasz2025meshsplats, zhou2024drivinggaussian}, and large-scale scene reconstruction~\cite{lin2024vastgaussian, liu2024citygaussian, xie2024gaussiancity, liu2024citygaussianv2}.

However, due to their approximation-based nature, 3D Gaussian methods often face challenges like high memory consumption, blurriness, and geometric inaccuracies. To address these limitations, recent efforts have focused on ``primitive-level modifications." Methods like GES~\cite{hamdi2024ges} generalize density distributions to reduce point counts, while others employ sharp kernels (e.g., clipped Gaussian~\cite{li20243d}, radial~\cite{huang2024deformable}, or convex kernels~\cite{held20243d}) to capture high-frequency appearance details. Assigning texture maps to planar primitives has also enabled more efficient scene representations~\cite{rong2024gstex,svitov2024billboard,song2024hdgstextured2dgaussian, chao2025texturedgaussians, weiss2024gaussian}. Despite these advances, such modifications mainly target \textit{high-quality appearance reconstruction}, leaving \textit{detailed geometry reconstruction} largely unaddressed. Several approaches adopt alternative kernel shapes, such as flat ellipsoids or planar disks (\emph{e.g.}, surfels)~\cite{DSS,huang20242d, dai2024high} to improve surface alignment. Yet, as a first-order geometric approximation, planar disks lack sufficient expressiveness; For example, highly curved surfaces requires dense primitive sampling, creating memory bottlenecks and rendering latency. 

In this paper, we extend the representations of ellipsoid~\cite{kerbl3Dgaussians} and elliptical disks to a more general quadratic framework. Specifically, we define primitive boundaries using an implicit function $f(x,y,z)=0$ where $f(x,y,z)<0$ for points inside the primitive and $f(x,y,z)>0$ otherwise. By adopting a quadratic form $f(x,y,z) = (x,y,z,1)^{\top}\mathbf{Q}(x,y,z,1) = 0$, where $\mathbf{Q}\in\mathbb{R}^{4\times4}$ is a conic matrix, we allow the representation of various shapes, including cylinder , ellipsoids, paraboloids, and hyperboloids. 
As a result, it generalizes 3D Gaussian primitives~\cite{kerbl3Dgaussians} and includes 2D Gaussian splatting~\cite{huang20242d} as a special case, allowing higher-order surface adaptation while incorporating kernel-level innovations such as shape kernels~\cite{huang2024deformable}, texture billboards~\cite{weiss2024gaussian,rong2024gstex,svitov2024billboard}, and weight distributions~\cite{hamdi2024ges}. 
In particular, there is an efficient ray-quadratic intersection algorithm~\cite{sigg2006gpu} that makes this representation practical and versatile in the context of splatting. Since the primitive is essentially a second-order primitive, we term our method Quadratic Gaussian Splatting (QGS).

However, for deformable quadratic primitives, defining a continuous density distribution is a key challenge. 
Unlike methods using rigid primitives (e.g., triangles, spheres) with Euclidean-based density, our deformable quadratic surfaces require a geometry-aware density metric. 
For instance, as Fig~\ref{fig:illustration} shows, when a primitive deforms from a disk to a paraboloid, Euclidean distance would measure straight-line distance, which ignores the curved surface. While Geodesic distance follows the surface, yields more accurate density modeling, it is generally difficult to compute.

\begin{figure}[t!]
    \centering
    \includegraphics[width=1.0\linewidth]{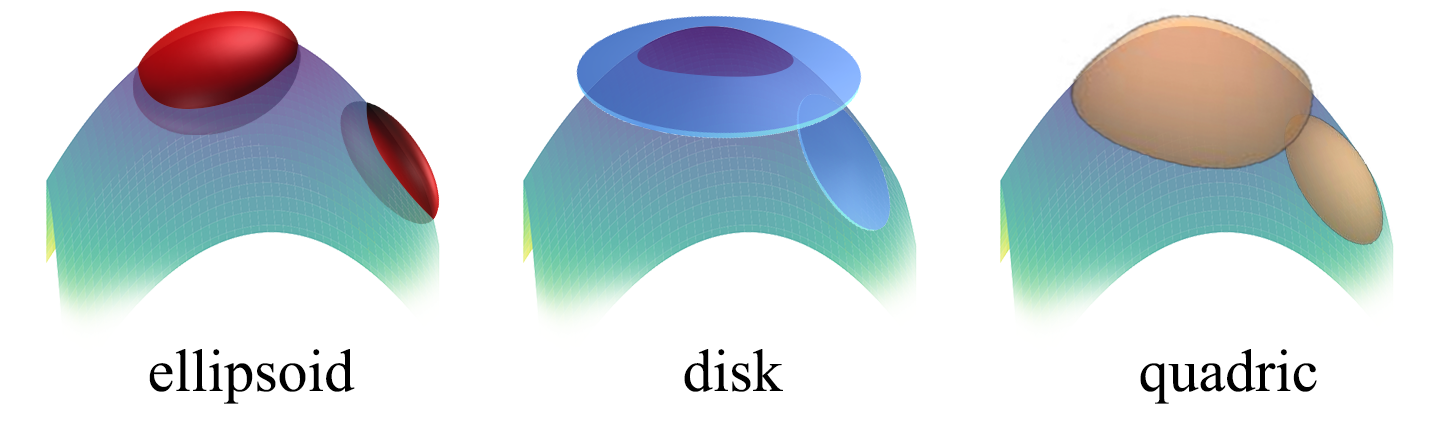}
    \\
    \begin{minipage}{0.32\linewidth}
        \includegraphics[width=1\linewidth]{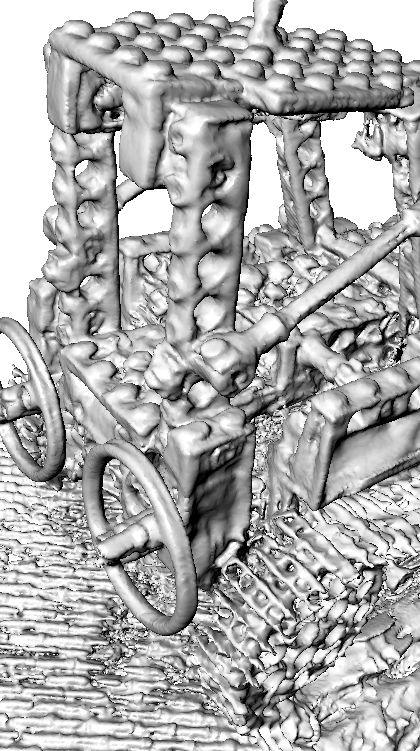}
        \vspace{-15pt}
        \caption*{3DGS}
    \end{minipage}
    \begin{minipage}{0.32\linewidth}
        \includegraphics[width=1\linewidth]{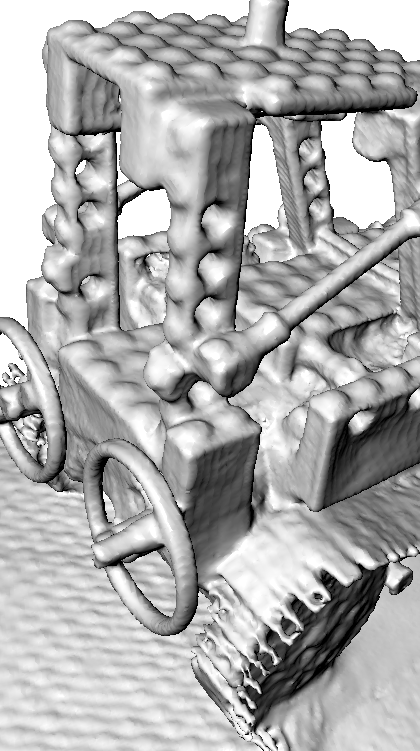}
        \vspace{-15pt}
        \caption*{2DGS}
    \end{minipage}
    \begin{minipage}{0.32\linewidth}
        \includegraphics[width=1\linewidth]{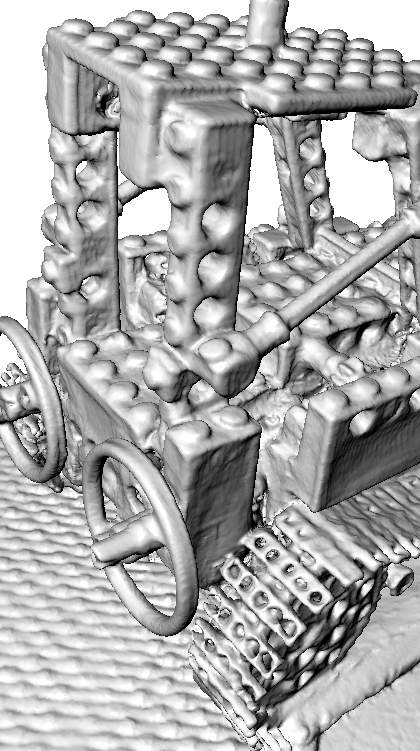}
        \vspace{-15pt}
        \caption*{QGS}
    \end{minipage}
    \vspace{-5pt}
    \caption{Demo and mesh reconstruction comparison on the Mip-NeRF 360 dataset. 
    ellipsoids struggle to align accurately with surfaces, producing coarse and incomplete meshes. Planes fail to capture high-curvature regions, resulting in oversmoothed reconstructions. In contrast, quadric surfaces effectively capture geometric edge regions, achieving detailed reconstruction results.
    }
    \label{fig:YYH}
    \vspace{-10pt}
\end{figure}

Luckily, we target at surface modeling so we constrain our quadratic representation to paraboloids—including planar ellipses as degenerate cases—where efficient closed-form geodesic solvers exist. 
This formulation maintains point-based rendering efficiency while enabling adaptive curvature modeling, where a single paraboloid approximates complex geometry that otherwise requires multiple planar surfels, achieving higher fidelity with similar primitive counts.
With such surfel-based representation and inspired by 2DGS~\cite{huang20242d}, 
we adopt a ray-splat intersection~\cite{sigg2006gpu} for splatting, maintaining multi-view consistency and perspective correctness. The surfel nature also allow QGS to leverage advanced splatting techniques that address perspective distortion~\cite{huang20242d} and popping effects \cite{radl2024stopthepop}, or advanced surface regularization~\cite{huang20242d, barron2022mip, fu2022geo} that improves overall geometry accuracy, albeit with added computational cost.

To validate the effectiveness, we evaluate QGS on three benchmark datasets—MipNeRF360~\cite{barron2022mip}, DTU~\cite{jensen2014large}, and Tanks and Temples~\cite{knapitsch2017tanks}—and demonstrate state-of-the-art surface reconstruction accuracy. Our ablation studies reveal that deformable primitives improve geometric fidelity by 33\% over 2DGS~\cite{huang20242d}, establishing QGS as a robust alternative for applications demanding higher geometric precision. We summarize our key contributions as follows:
\begin{itemize}
\item A quadric surfel representation that generalizes 2D Gaussians to higher-order surfaces (e.g., paraboloids), enabling adaptive curvature modeling and capturing high frequency details.
\item Geodesic-aware density distributions, the first method to unify deformable primitives with surface-aligned density modeling via closed-form geodesic solvers.
\item State-of-the-art reconstruction: QGS reduces geometric error by 33\% over 2DGS and 27\% over GOF on DTU~\cite{jensen2014large} while maintaining competitive training and rendering speeds.
\end{itemize}

%% file: sec/2_relatedwork.tex
\section{Related work}
\label{sec:related_work}
\subsection{Novel View Synthesis}
Novel view synthesis has advanced rapidly, particularly since NeRF \cite{mildenhall2021nerf}. 
NeRF represents scene geometry and view-dependent appearance using a multi-layer perceptron (MLP) and optimizes it end-to-end through volumetric rendering to generate realistic images. 
Subsequent methods have introduced various enhancements. 
Mip-NeRF \cite{barron2021mip}, Mip-NeRF 360 \cite{barron2022mip}, and Zip-NeRF \cite{barron2023zip} addressed aliasing issues by refining sampling strategies, while I-NGP \cite{muller2022instant} and DVGO \cite{sun2022direct} significantly improved training and rendering speeds by replacing MLPs with feature grids.\par
More recently, 3DGS \cite{kerbl3Dgaussians} has gained attention for its superior rendering quality and real-time performance, often surpassing NeRF \cite{yu2024mip,radl2024stopthepop,lu2024scaffold,zhou2024drivinggaussian}. 
Mip-Splatting \cite{yu2024mip} introduced a smoothing filter to reduce high-frequency artifacts, while StopThePop \cite{radl2024stopthepop} improved multi-view consistency through per-tile and per-pixel sorting. 
Scaffold-GS \cite{lu2024scaffold} optimized Gaussian distributions by reducing redundancy and improving rendering quality through anchor-based 3D Gaussian placement.
\subsection{Neural Surface Reconstruction}
Neural volumetric rendering methods often produce smoother and more complete reconstructions than traditional approaches. 
NeuS \cite{wang2021neus} and VolSDF \cite{yariv2021volume} introduced signed distance fields (SDF) to model scene geometry within the NeRF framework \cite{mildenhall2021nerf}. 
Geo-NeuS \cite{fu2022geo} and NeuralWarp \cite{grabocka2018neuralwarp} further improved reconstruction by enforcing multi-view consistency. 
Neuralangelo \cite{li2023neuralangelo} reduced reliance on MLPs by incorporating feature grids and numerical gradients, enhancing geometric fidelity.\par
However, neural rendering-based surface reconstruction often takes hours to converge using only image supervision. 
Additionally, the structured implicit representations make direct access and editing of scene geometry challenging.
\subsection{Gaussian Splatting Surface Reconstruction}
With the rapid advancement of 3DGS across various domains, Gaussian splatting methods have also significantly improved surface reconstruction. 
SuGaR \cite{guedon2024sugar} introduced regularization terms to encourage Gaussians to align with scene surfaces, constructing volumetric density fields from Gaussian ellipsoids and applying Poisson reconstruction \cite{kazhdan2006poisson} to extract meshes. 
GSDF \cite{yu2024gsdf} and NeuSG \cite{chen2023neusg} combined 3D Gaussian ellipsoids with signed distance fields (SDFs), achieving high-quality geometric reconstructions and rendering results.
Other approaches have leveraged ray-Gaussian intersections for improved accuracy. 
Rade-GS \cite{zhang2024rade}, GOF \cite{yu2024gaussian}, and PGSR \cite{chen2024pgsr} computed ray-ellipsoid intersections to obtain unbiased depth estimates and applied normal consistency supervision from 2DGS \cite{huang20242d} to achieve state-of-the-art reconstruction. 
2DGS further refined this approach by flattening Gaussian ellipsoids into disks, better aligning primitives with surfaces while introducing additional regularization losses for enhanced geometric constraints. 
MVG-splatting \cite{li2024mvg} extended 2DGS by incorporating multi-view consistency constraints, enabling more complex surface reconstructions. 
In this work, we propose Quadratic Gaussian Splatting (QGS), a novel surface representation based on quadric surfaces designed to improve the geometric fitting of primitives. 
We establish Gaussian distributions on quadrics, enabling end-to-end optimization. 
Additionally, by integrating multi-view consistency supervision, QGS achieves state-of-the-art geometric reconstruction and high-quality rendering.

%% file: sec/3_method.tex
\section{Method}
\label{sec:method}
\subsection{Preliminary}
Kerbl et al. \cite{kerbl3Dgaussians} proposed representing a scene using 3D Gaussian ellipsoids as primitives and render images using differentiable volume splatting. 

\begin{equation}
    C(\mathbf{p})=\sum_{i=0}^{N-1}G_i^{2D}(\mathbf{p})\alpha_ic_i\prod_{j=0}^{i-1}(1-G_j^{2D}(\mathbf{p})\alpha_j)
\end{equation}\par
Here, $\alpha_i$ represents the opacity, $c_i$ denotes the color of each Gaussian primitive modeled by spherical harmonics, which is also used in our method, and $\mathbf{p}$ is the pixel coordinate.
Finally, each Gaussian primitive is optimized by minimizing the photometric loss.\par
\noindent\textbf{GS mesh reconstruction. }
Vanilla 3DGS \cite{kerbl3Dgaussians} can render high-quality images, but it yields suboptimal results for scene geometry reconstruction. 
In subsequent surface reconstruction methods, volumetric approaches like GOF \cite{yu2024gaussian} and RadeGS \cite{zhang2024rade} employ ray-splat intersection techniques, achieving SOTA reconstruction quality but limiting the consistency of normal and depth.
In contrast, 2DGS \cite{huang20242d} defines the 2D Gaussian distribution within a planar disk, which inherently provides consistent normals and depths across multiple views.
But the disk is only a first-order approximation of the surface, which often leads to overly smooth reconstruction results in 2DGS, as shown in Fig~\ref{fig:YYH}.

\begin{figure}[t!]
    \centering
    \includegraphics[width=0.5\linewidth]{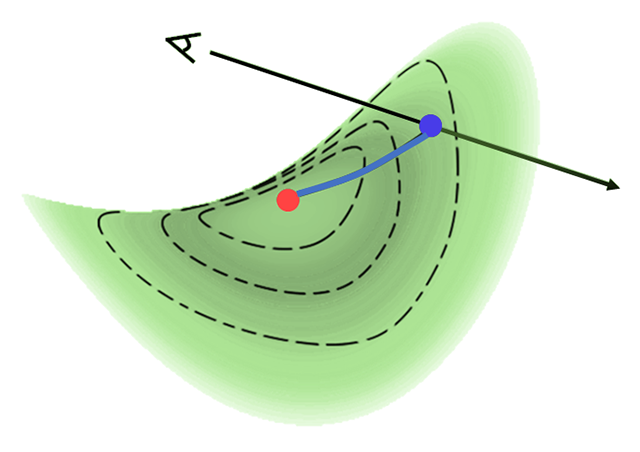}
    \vspace{-5pt}
    \caption{Illustration of a Gaussian on a quadric surface. The blue dot marks the Gaussian centroid, the red dot indicates the ray-splat intersection, and the red line represents the geodesic.}
    \label{fig:quadratic}
    \vspace{-15pt}
\end{figure}

\subsection{Quadratic Gaussian Splatting}
\label{QGS_main}
To enhance the geometric fitting capability of the surface representation, we present differentiable Quadratic Gaussian Splatting, as shown in Fig~\ref{fig:teaser}. 
We will first introduce the Quadric Gaussian Model, then discuss the splatting design for quadrics, and finally explain the optimization process.\par
\noindent\textbf{Quadratic Gaussian Model.}
Given a homogeneous coordinate $\mathbf{x} = [x, y, z, 1]^T \in \mathbb{R}^4$, a quadric surface can be defined as the solution set to the following equation:
\begin{equation}
    \begin{aligned}
f(x,y,z)&=Ax^2+2Bxy+2Cxz+2Dx+Ey^2\\&+2Fyz+2Gy+Hz^2+2Iz+J\\
&=\left[\begin{matrix}x&y&z&1\end{matrix}\right]\left[\begin{matrix}A&B&C&D\\B&E&F&G\\C&F&H&I\\D&G&I&J\end{matrix}\right]\left[\begin{matrix}x\\y\\z\\1\end{matrix}\right]\\
&=\mathbf{x}^T\mathbf{Q}\mathbf{x}=0
\end{aligned}
\end{equation}\par
We apply a congruent transformation to obtain the transformation from object space to parameter space.
\begin{equation}
\begin{aligned}
    \mathbf{Q}=\mathbf{T}^{-T} \mathbf{D} \mathbf{T}^{-1},
    \mathrm{with}\;&\mathbf{D}\;\mathrm{symmetrical}\\
    &\mathbf{T}=\left[\begin{matrix}\mathbf{R}&\mathbf{c}\\0&1\end{matrix}\right]
\end{aligned}
\end{equation}\par
Here, $\mathbf{c}$ denotes the position of the quadric, and $\mathbf{R} = [\mathbf{r}_1, \mathbf{r}_2, \mathbf{r}_3]$ denotes the orientation of the quadric in the object space. 
The matrix $\mathbf{D}$ defines the surface scale and shape: $\mathbf{D} = \mathrm{diag}(s_1^2, s_2^2, s_3^2, 1)$ yields an ellipsoid, while $\mathbf{D} = \mathrm{diag}(s_1^2, s_2^2, 0, 0)$ produces a plane.\par
\begin{figure}[t]
    \centering
    \includegraphics[width=\linewidth]{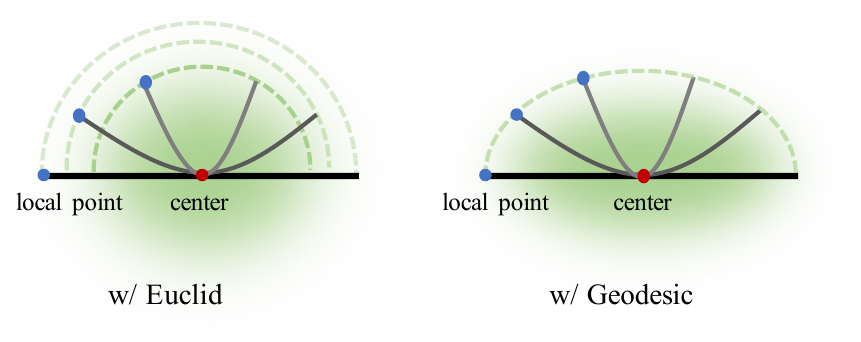}
    \vspace{-20pt}
    \caption{Density distribution among deformable primitives (disk→paraboloid). (a) Euclidean distance fails to adapt to curvature during primitive deformation, distorting density distributions (uneven weight in the blue point). (b) Our geodesic distance follows intrinsic curvature, maintaining spatial coherence across deformation and preventing uneven splatting artifacts.}
    \label{fig:illustration}
    \vspace{-10pt}
\end{figure}
\begin{figure}[!t]
    \centering
    \begin{minipage}{0.45\linewidth}
        \includegraphics[width=1\linewidth]{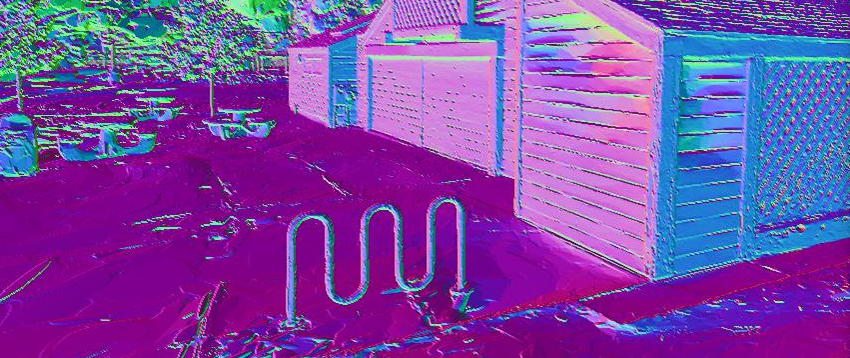}
        \vspace{-15pt}
        \caption*{(a) w/ Euclid}
    \end{minipage}
    \begin{minipage}{0.45\linewidth}
    \centering
        \includegraphics[width=1\linewidth]{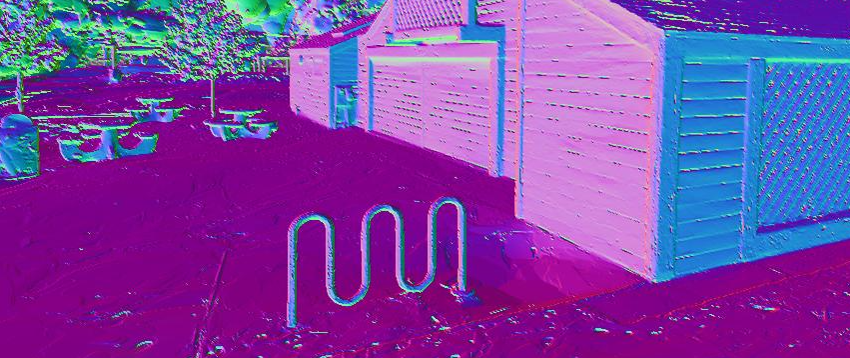}
        \vspace{-15pt}
        \caption*{(b) w/ Geodesic}
    \end{minipage}
    \vspace{-5pt}
    \caption{Comparison between Euclidean and geodesic distances. The first row presents schematic illustrations, while the second row shows normal comparisons. The left column uses Euclidean distance, while the right column employs geodesic distance.}
    \label{fig:Euclid_Geodesic}
    \vspace{-8pt}
\end{figure}

To compute the Gaussian weight at any surface point, we first define a measure on the surface to establish the Gaussian distribution. 
A straightforward approach is to construct the Gaussian distribution using Euclidean distance. 
However, this leads to inconsistencies when the surface undergoes deformation, resulting in uneven splatting  reconstruction artifacts, as shown in Fig. \ref{fig:Euclid_Geodesic}.
To address this, we use geodesic distance to ensure that weights adapt consistently to surface deformations. Fig.~\ref{fig:illustration} shows a 1D illustration. \par
However, not all geodesics have closed-form solutions. 
When $\mathbf{Q}$ is an ellipsoid or hyperboloid, computing the geodesic length typically requires numerical methods. Luckily, as we aim at surface reconstruction, we focus on paraboloid with the following form:
\begin{equation}
\begin{aligned}
f(x,y,z)
&=\mathbf{x}^T\left[\begin{matrix}\mathbf{R}&\mathbf{c}\\0&1\end{matrix}\right]^{-T}\mathbf{D}\left[\begin{matrix}\mathbf{R}&\mathbf{c}\\0&1\end{matrix}\right]^{-1}\mathbf{x}\\
&=\hat{\mathbf{x}}^T\left[\begin{matrix}\frac{d_{11}}{s_1^2}&0&0&0\\0&\frac{d_{22}}{s_2^2}&0&0\\0&0&0&-\frac{d_{33}}{2s_3}\\0&0&-\frac{d_{33}}{2s_3}&0\end{matrix}\right]\hat{\mathbf{x}}\\
&=\frac{d_{11}}{s_1^2}\hat{x}^2+\frac{d_{22}}{s_2^2}\hat{y}^2-\frac{d_{33}}{s_3}\hat{z}=0
\end{aligned}
\end{equation}\par

Here and henceforth, we use $\hat{\cdot}$ to denote Gaussian local coordinate.
$d_{ii} \in \{0, \pm 1\}$ determines whether the paraboloid is elliptic, hyperbolic, or planar. 
However, since $d_{ii}$ is discrete, the primitive cannot transition smoothly between elliptic and hyperbolic paraboloids. 
To resolve this, we introduce a signed scale for a differentiable transition between paraboloid types.
\begin{equation}
    f(\hat{x},\hat{y},\hat{z})=\frac{\mathrm{sign}(s_1)}{s_1^2}\hat{x}^2+\frac{\mathrm{sign}(s_2)}{s_2^2}\hat{y}^2-\frac{1}{s_3}\hat{z}=0
    \label{eq:implicit_function}
\end{equation}\par

In vanilla 3DGS \cite{kerbl3Dgaussians}, the scale is obtained through the $\exp$ activation function, i.e., $s(x) = \exp(x)$. To introduce a sign, we add another variable $t$ to control the sign, i.e., $s(x, t) = \tanh(t)\exp(x)$.

\noindent\textbf{Gaussian Distribution on Quadric.}  
We will now describe how to define a Gaussian distribution on a paraboloid using geodesic lines.
First, paraboloid (Equation \ref{eq:implicit_function}) could be expressed in explicit form:
\begin{equation}
    \hat{z}(\hat{x},\hat{y})=s_3(\frac{\mathrm{sign}(s_1)}{s_1^2}\hat{x}^2+\frac{\mathrm{sign}(s_2)}{s_2^2}\hat{y}^2)
    \label{eq:explicit_function}
\end{equation}\par
We convert it isometrically into cylindrical coordinates, i.e. $\hat{x}=\rho\cos\theta,\hat{y}=\rho\sin\theta$. And we rewrite Equation \ref{eq:explicit_function} as:
\begin{equation}
\begin{split}
    \hat{z}(\theta,\rho)=\;&s_3(\frac{\mathrm{sign}(s_1)\cos^2\theta}{s_1^2}+\frac{\mathrm{sign}(s_2)\sin^2\theta}{s_2^2})\rho^2\\
    =\;&a(\theta)\rho^2
\end{split}
\end{equation}\par
Due to the paraboloid's symmetry, for any point $\hat{\mathbf{p}}_0 = (\rho_0, \theta_0, \hat{z}(\theta_0, \rho_0))$, the intersection line of the plane $\theta = \theta_0$ with the paraboloid: $\hat{z}(\theta_0, \rho), \rho \in (0, \rho_0)$, is the geodesic to the origin. 
Then the geodesic distance is the arc length $l$ of this curve, as shown in Fig \ref{fig:quadratic}.

\begin{equation}
    \begin{aligned}
l(a,\rho_0)&=\int_0^{\rho_0}\sqrt{1+(2at)^2}\mathrm{d}t\\
&=\frac{\ln(\sqrt{u^2+1}+u)+u\sqrt{u^2+1}}{4a}\\
&\mathrm{where}\;u=2a\rho_0
\end{aligned}
\label{eq:geodesic}
\end{equation}\par
For the derivation of the integral, please refer to the supplementary materials. 
We then define the mean of the 2D Gaussian distribution on the surface at the origin of the quadric surface, with $(s_1, s_2)$ representing the principal axis variances of the Gaussian.
Since the contours of the 2D Gaussian distribution form an ellipse:
\begin{equation}
    \frac{\rho^2\cos^2\theta}{s_1^2}+\frac{\rho^2\sin^2\theta}{s_2^2}=1
\end{equation}\par
Given a point $(\theta_0, \rho_0)$ on the ellipse, $\rho_0$ represents the standard deviation of the 2D Gaussian distribution in the $\theta_0$ direction.
\begin{equation}
    \sigma(\theta_0)=\rho_0=\frac{s_1s_2}{\sqrt{(s_2\cos\theta_0)^2+(s_1\sin\theta_0)^2}}
\end{equation}

Thus, for any point $\hat{\mathbf{p}}_0$ on the surface, we can define the corresponding Gaussian function value as: 
\begin{equation}
    G(\hat{\mathbf{p}}_0(\theta_0,\rho_0))=\exp(-\frac{(l(a(\theta_0),\rho_0))^2}{2(\sigma(\theta_0))^2})
\end{equation}\par
Notably, when $|s_3| \rightarrow 0$, the paraboloid becomes equivalent to a disk.   
Furthermore, as $x \rightarrow 0$, we have $\sqrt{1+x} \rightarrow 1$ and $\ln(1+x) \sim x$. Thus, as $|s_3| \rightarrow 0$, we get $a \rightarrow 0$ and $l \rightarrow \rho_0$ by Equation \ref{eq:geodesic}, meaning the geodesic distance becomes equivalent to the Euclidean distance.
This indicates that 2DGS can be regarded as a specific degenerate case of QGS, whose more generalized nature allows it to fit high-curvature regions effectively.

\begin{figure}[!t]
    \centering
    \begin{minipage}{0.45\linewidth}
        \includegraphics[width=1\linewidth]{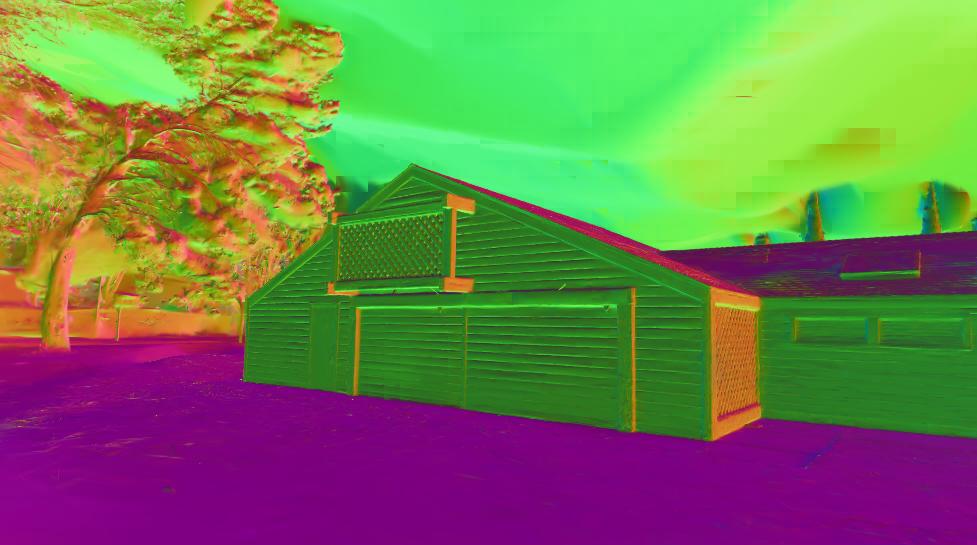}
        \vspace{-15pt}
        \caption*{(a) normal}
    \end{minipage}
    \begin{minipage}{0.45\linewidth}
        \includegraphics[width=1\linewidth]{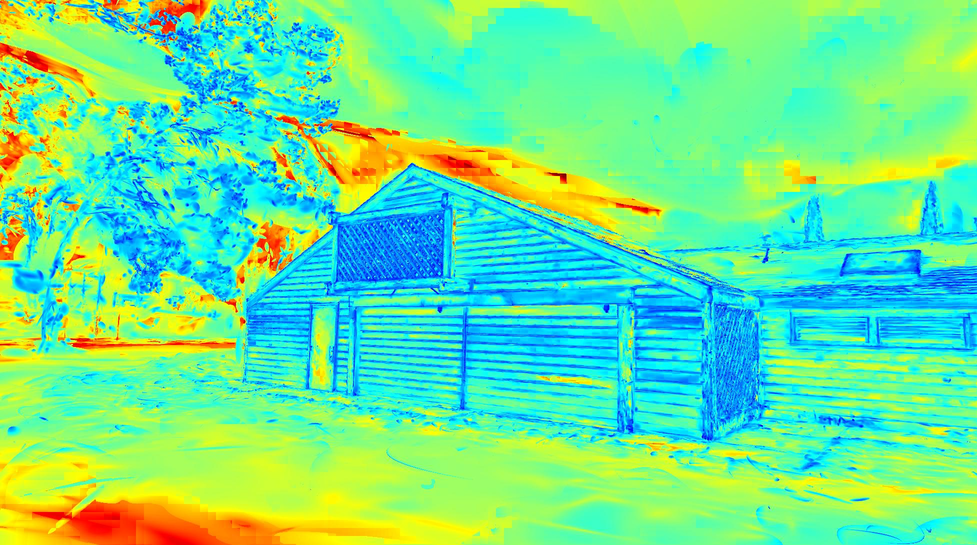}
        \vspace{-15pt}
        \caption*{(b) curvature}
    \end{minipage}
    \vspace{-5pt}
    \caption{The (a) normal map and (b) curvature map of a scene. In the curvature map, blue indicates higher curvature and curved areas, while red indicates lower curvature and flatter areas.}
    \label{fig:normal_and_curvature}
    \vspace{-8pt}
\end{figure}

\subsubsection{Splatting}
Although Sigg et al. \cite{sigg2006gpu} derived the Ray-Quadric Intersection, QGS integrates Gaussian distribution, necessitating a redefinition of the intersection for multi-view consistency.\par
\noindent\textbf{Ray-splat Intersection.} 
Let the camera center in the Gaussian local space be denoted as $\hat{\mathbf{o}} \in \mathbb{R}^{3 \times 1}$ and the ray direction as $\hat{\mathbf{d}} \in \mathbb{R}^{3 \times 1}$. 
A point on the ray can be defined as $\hat{\mathbf{p}} = \hat{\mathbf{o}} + t\hat{\mathbf{d}}$. 
By substituting $\hat{\mathbf{p}}$ into the Equation \ref{eq:explicit_function}, we solve a quadratic equation to obtain two intersection points: the nearer point $\hat{\mathbf{p}}_n = (\hat{x}_n, \hat{y}_n, \hat{z}_n)$ and the farther point $\hat{\mathbf{p}}_f = (\hat{x}_f, \hat{y}_f, \hat{z}_f)$, with $t_n \leq t_f$. 
To ensure multi-view consistency in QGS, only one valid point of the two is selected. 
First, if the geodesic distance of $\hat{\mathbf{p}}_n$ is within $3\sigma(\theta_n)$, we choose $\hat{\mathbf{p}}_n$. 
If not, we check if $\hat{\mathbf{p}}_f$ is within $3\sigma(\theta_f)$. If so, we select $\hat{\mathbf{p}}_f$. 
If neither condition is met, we assume no intersection between the ray $\hat{\mathbf{p}}(t)$ and the primitive. 
This assumption is generally valid because the significant weight typically causes the first point to occlude the second point. 
The derivation is provided in the supplementary material. \par
\noindent\textbf{Normal and Curvature.} 
Similar to 2DGS \cite{huang20242d}, QGS is a surface-based representation that naturally possesses multi-view consistent geometric properties, making it straightforward to compute surface normals. Given any point $\hat{\mathbf{p}}_0 = (\hat{x}_0, \hat{y}_0, \hat{z}(\hat{x}_0, \hat{y}_0))$ on the surface, we can take the partial derivatives of the Equation \ref{eq:implicit_function}, yielding:
\begin{equation}
    \hat{\mathbf{n}}_0(\hat{\mathbf{p}}_0)=(\frac{2\mathrm{sign}(s_1)}{s_1^2}\hat{x}_0,\frac{2\mathrm{sign}(s_2)}{s_2^2}\hat{y}_0,-\frac{1}{s_3})
\end{equation}\par
As QGS provides a second-order fit, it naturally outputs second-order geometric information like curvature, which describes surface bending.
For each QGS primitive, the Gaussian curvature at the ray-splat intersection can be computed analytically, with detailed derivation available in the supplementary material. 
Let $\lambda_1 = \mathrm{sign}(s_1) \cdot s_3 / s_1^2$ and $\lambda_2 = \mathrm{sign}(s_2) \cdot s_3 / s_2^2$. The curvature at point $\hat{\mathbf{p}}_0$ is:
\begin{equation}
    \hat{K}_0(\hat{\mathbf{p}}_0)=\frac{4\lambda_1\lambda_2}{(1+4\lambda_1^2\hat{x}_0^2+4\lambda_2^2\hat{y}_0^2)^2}
\end{equation}\par
We render the normal map $\mathbf{N}$ and curvature map $\mathbf{K}$ by Equation \ref{eq:NK_render} for a given viewpoint using alpha-blending, as shown in the Fig \ref{fig:normal_and_curvature}.

\begin{equation}
\begin{aligned}
    \{\mathbf{N}, \mathbf{K}\}&=\sum_{i=0}^{N-1}G_i\alpha_i\prod_{j=0}^{i-1}(1-G_j\alpha_j)\{\textbf{n}_i, K_i\}\\
\end{aligned}
\label{eq:NK_render}
\end{equation}
\par

\begin{figure}[!t]
    \centering
    \begin{minipage}{0.40\linewidth}
        \includegraphics[width=1\linewidth]{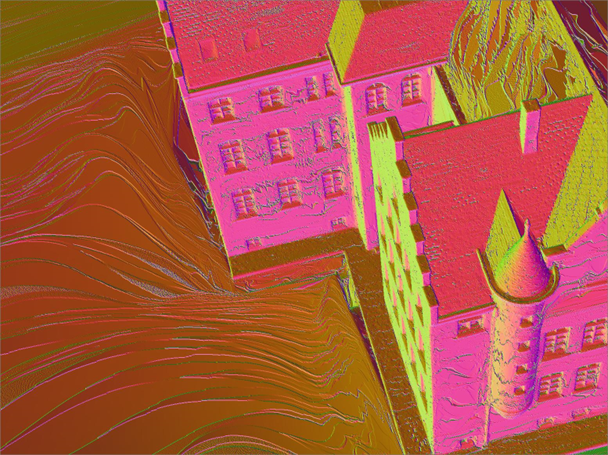}
        \vspace{-15pt}
        \caption*{(a) w/o~resorting}
    \end{minipage}
    \begin{minipage}{0.40\linewidth}
        \includegraphics[width=1\linewidth]{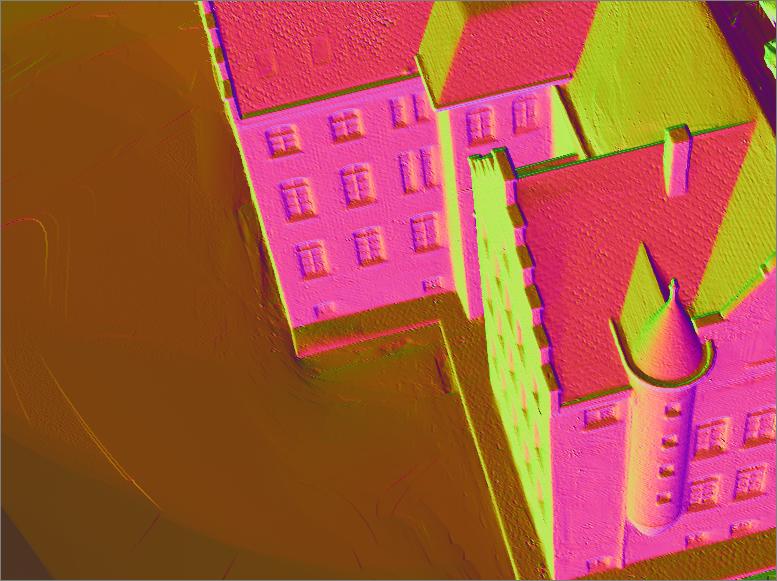}
        \vspace{-15pt}
        \caption*{(b) w/~resorting}
    \end{minipage}
    \vspace{-5pt}
    \caption{Comparison of depth normals with and without resorting. Centroid-only sorting (a) introduces stripe artifacts, while resorting (b) produces smoother and more consistent normals.}
    \label{fig:comp_PR}
    \vspace{-10pt}
\end{figure}
\noindent{\textbf{Per-pixel Resorting.~}}We adopt the per-pixel resorting method from Stopthepop to address the depth pop-out issue, as illustrated in Fig~\ref{fig:comp_PR}. Implementation and derivation details are provided in the supplementary material.

\subsection{Optimization}
We now introduce the our training objectives during optimization.\par
\noindent\textbf{Depth Distortion. } 
We adopt the depth distortion loss and normal consistency loss proposed by 2DGS \cite{huang20242d}.
\begin{equation}
    \mathcal{L}_d=\sum_{i=0}^{N-1}\sum_{j=0}^{i-1}\omega_i\omega_j(t_i-t_j)^2
\end{equation}\par
Here, $\omega_i = \bar{\alpha}_i T_i$ denotes the alpha-blending weight of the i-th Gaussian, and $t_i$ represents the depth at the ray-splat intersection. 
$i,j$ index the Gaussians along the ray.\par
\noindent\textbf{Curvature Guided Normal Consistency. }
2DGS \cite{huang20242d} introduces the normal consistency loss to ensure that all primitives on a ray are locally aligned with the actual surface.
\begin{equation}
\begin{aligned}
    \mathcal{L}_n&=\sum_i\omega_i(1-\mathbf{n}_i^T\mathbf{N})\\
\end{aligned}
\end{equation}\par
Here, $\mathbf{n}_i$ represents the splat normal facing the camera, while $\mathbf{N}$ is computed by differentiating the depth point $\mathbf{p}$ from neighboring pixels. 
However, neighboring pixels may violate the local planar assumption, particularly in edge regions. 
Using normal consistency loss in edge areas can introduce errors, contributing to the over-smoothing observed in 2DGS. 
A straightforward approach is to approximate geometric edges using image edges, which can then guide normal consistency supervision \cite{chen2024pgsr,li2024mvg}. 
However, we found that, in most scenes, image edges do not fully correspond to geometric edges, especially in uniformly lit areas.
Thus, we use the curvature map, which more accurately corresponds to geometric edges and is efficiently and uniquely generated by QGS, to guide normal supervision.\par
\vspace{-10pt}
\begin{equation}
\begin{aligned}
    \lambda_K(K(u,v))=\;&1-\mathrm{sigmoid}(\ln(|K(u,v)|+\varepsilon)\\
\mathcal{L}_{Kn}(u,v)&=\lambda_K(K(u,v))\mathcal{L}_n(u,v)
\end{aligned}
\end{equation}
\noindent\textbf{Multi-view regularization Loss. }To ensure a fairer comparison with plane-based PGSR \cite{chen2024pgsr}, we incorporate PGSR's multi-view regularization, computing warp and photometric consistency losses via homography transformations.
\begin{equation}
\begin{aligned}
    \mathbf{H}_{rn}=&\mathbf{K}_{n}(\mathbf{R}_{rn}+\frac{\mathbf{T}_{rn}\mathbf{n}_r^T}{\mathbf{n}_r^T\mathbf{X}_r})\mathbf{K}_r^{-1}\\
    \mathcal{L}_{Mv}=&\frac{1}{V}\sum_{\mathbf{p}_r\in V}||\mathbf{p}_r-\mathbf{H}_{nr}\mathbf{H}_{rn}\mathbf{p}_r||+\\
    &(1-NCC(\mathbf{I}_r(\mathbf{p}_r),\mathbf{I}_n(\mathbf{H}_{rn}\mathbf{p}_r)))
\end{aligned}
\end{equation}
Here, $\mathbf{n}_r$ represents the normal map, and $\mathbf{X}_r$ denotes the 3D points projected from the depth map. 
$\mathbf{R}_{rn}$ represents the relative rotation, 
$\mathbf{K}$ is the camera intrinsic, $\mathbf{p}_r$ denotes the pixel coordinates, and $\mathbf{I}$ refers to the image intensity. 
This loss enforces multi-view constraints by computing the reprojection error from the depth and normal maps, along with normalized cross-correlation (NCC) \cite{yoo2009fast}.\par
\noindent\textbf{Final Loss.} 
Finally, we input a sparse point cloud and posed images to optimize QGS with the following loss function:
\begin{equation}
    \mathcal{L}=\mathcal{L}_c+\lambda_d\mathcal{L}_{d}+\lambda_n\mathcal{L}_{Kn}+\lambda_{Mv}\mathcal{L}_{Mv}
\end{equation}

%% file: sec/4_experiments.tex
\section{Experiment}
\begin{figure*}[!t]
    \centering
    \begin{minipage}{0.05\linewidth}
        \centering
        \rotatebox{90}{\textbf{scan24}}
    \end{minipage}
    \begin{minipage}{0.224\linewidth}
        \includegraphics[width=\linewidth]{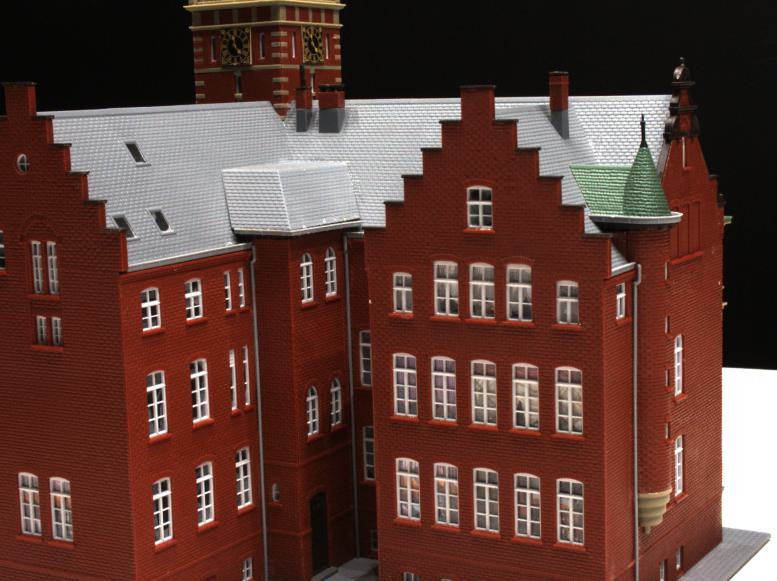}
    \end{minipage}
    \begin{minipage}{0.15652\linewidth}
    \begin{tikzpicture}
            \node[anchor=south west,inner sep=0] (image) at (0,0)
                {\includegraphics[width=\linewidth]{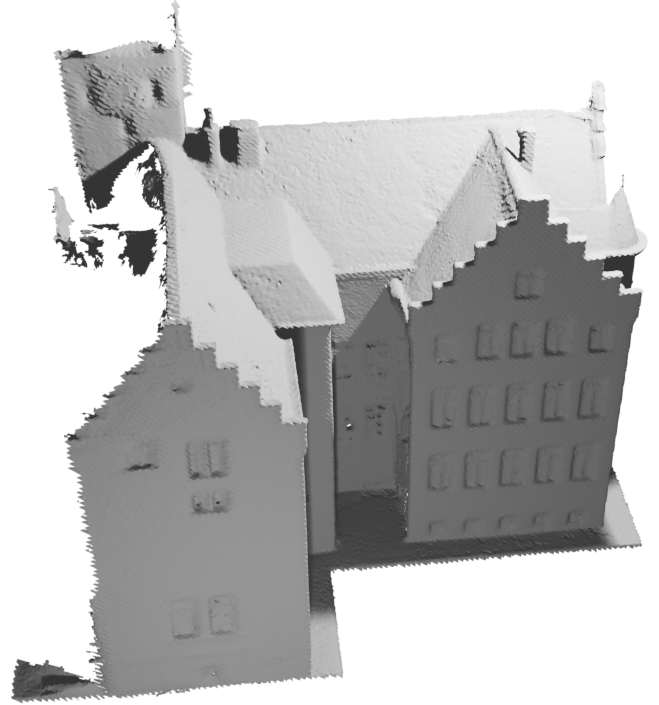}};
            
            \node[anchor=south west] at (1.44, 0.0) {\includegraphics[width=0.4\linewidth]{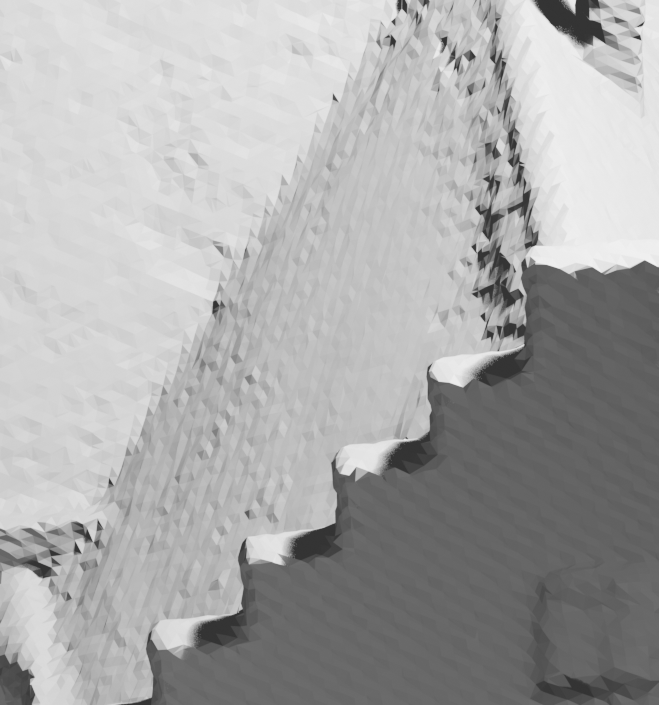}};
            \begin{scope}[x={(image.south east)},y={(image.north west)}]
                \draw[red, thick] (0.57,0.045) rectangle (0.97,0.44);
            \end{scope}
        \end{tikzpicture}
    \end{minipage}
    \begin{minipage}{0.15652\linewidth}
        \begin{tikzpicture}
            \node[anchor=south west,inner sep=0] (image) at (0,0)
                {\includegraphics[width=\linewidth]{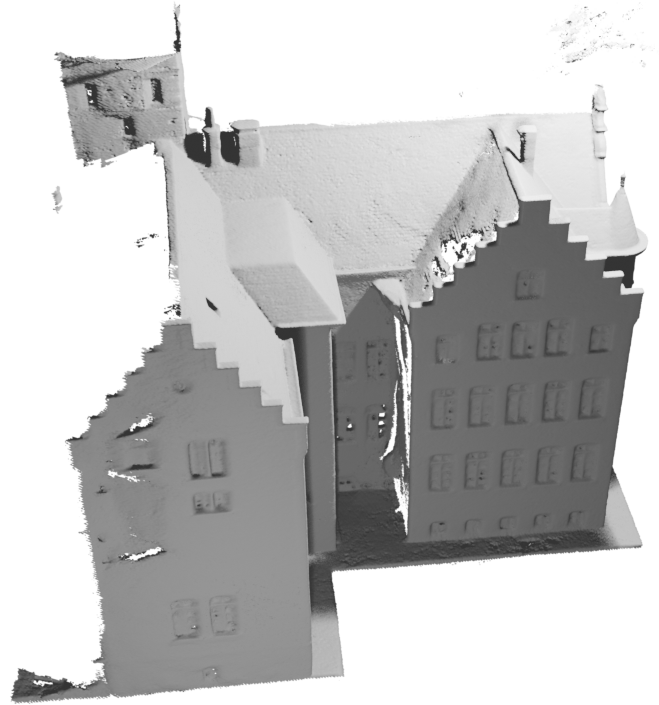}};
            
            \node[anchor=south west] at (1.44, 0.0) {\includegraphics[width=0.4\linewidth]{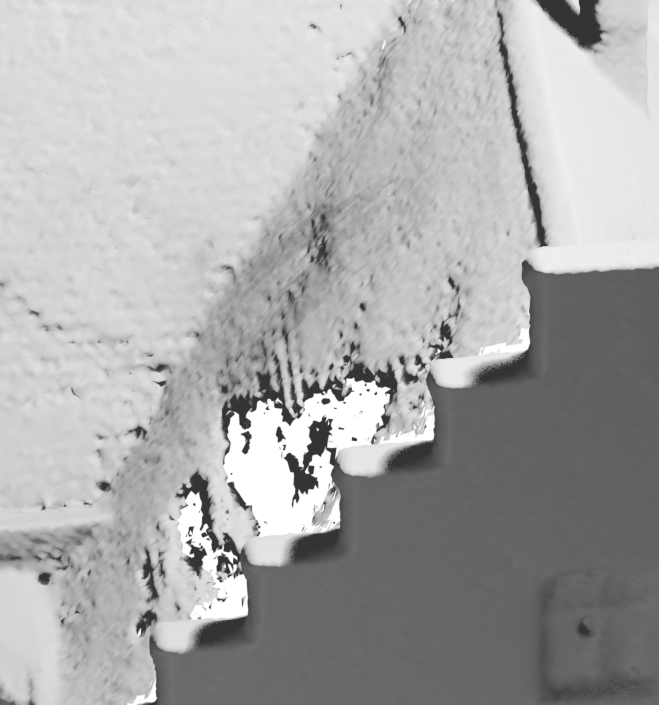}};
            \begin{scope}[x={(image.south east)},y={(image.north west)}]
                \draw[red, thick] (0.57,0.045) rectangle (0.97,0.44);
            \end{scope}
        \end{tikzpicture}
    \end{minipage}
    \begin{minipage}{0.15652\linewidth}
        \begin{tikzpicture}
            \node[anchor=south west,inner sep=0] (image) at (0,0)
                {\includegraphics[width=\linewidth]{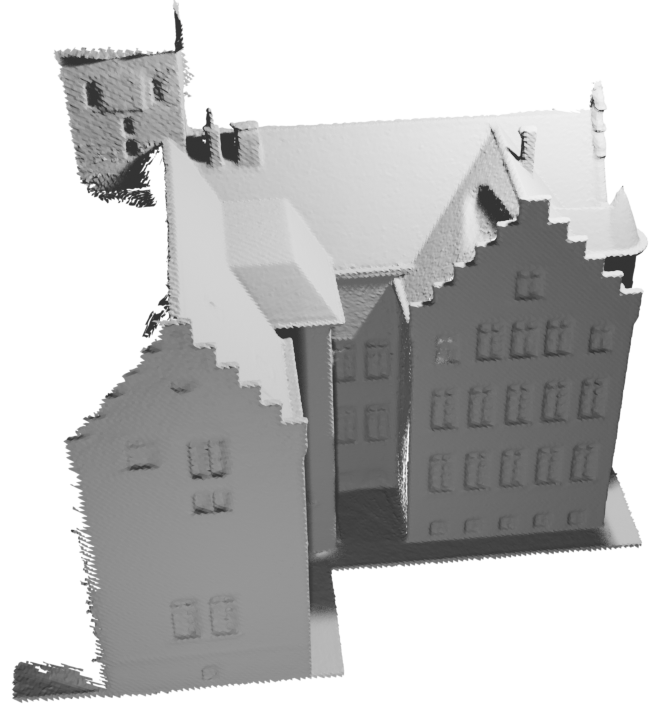}};
            
            \node[anchor=south west] at (1.44, 0.0) {\includegraphics[width=0.4\linewidth]{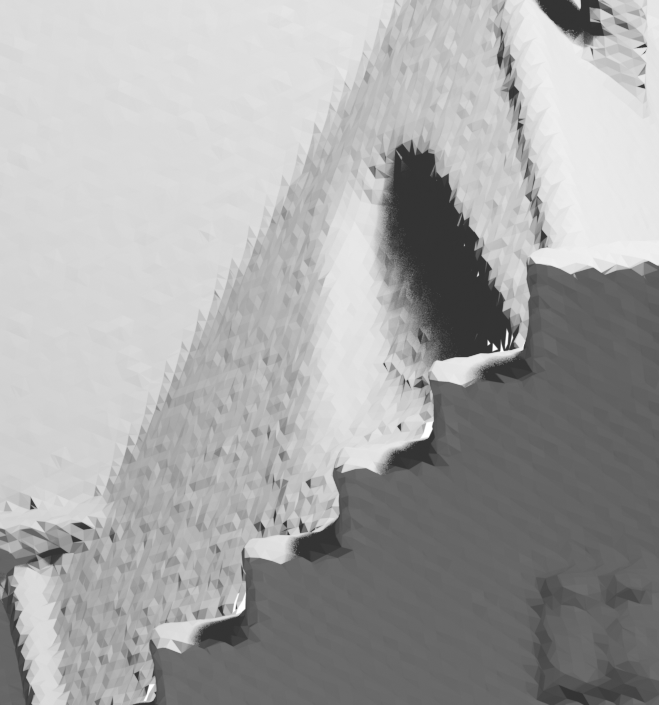}};
            \begin{scope}[x={(image.south east)},y={(image.north west)}]
                \draw[red, thick] (0.57,0.045) rectangle (0.97,0.44);
            \end{scope}
        \end{tikzpicture}
    \end{minipage}
    \begin{minipage}{0.15652\linewidth}
        \begin{tikzpicture}
            \node[anchor=south west,inner sep=0] (image) at (0,0)
                {\includegraphics[width=\linewidth]{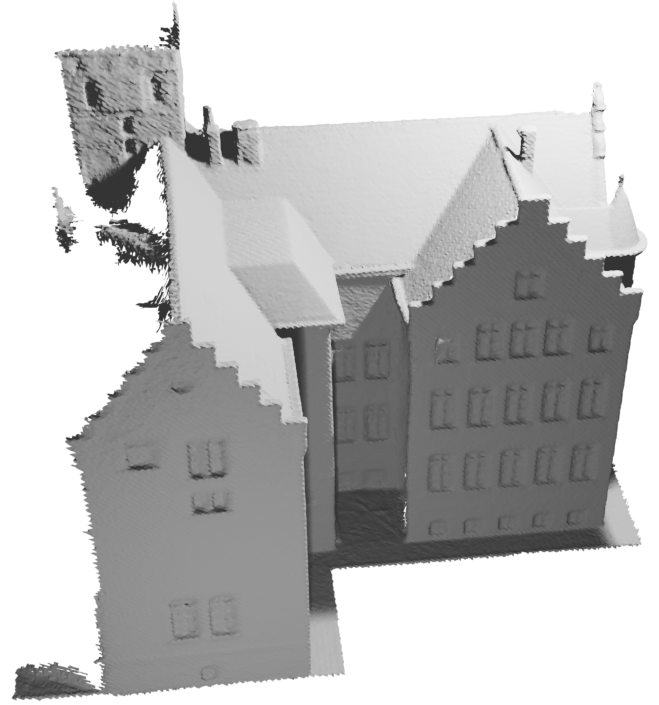}};
            
            \node[anchor=south west] at (1.44, 0.0) {\includegraphics[width=0.4\linewidth]{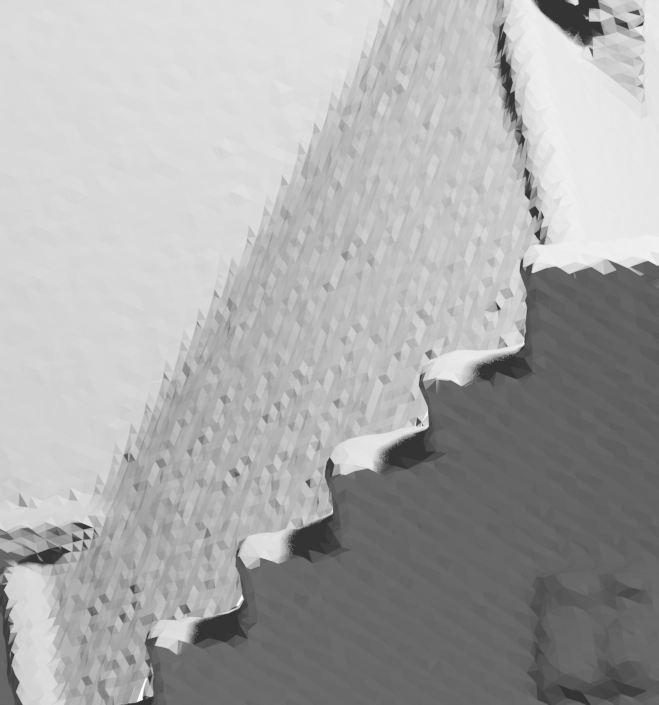}};
            \begin{scope}[x={(image.south east)},y={(image.north west)}]
                \draw[red, thick] (0.57,0.045) rectangle (0.97,0.44);
            \end{scope}
        \end{tikzpicture}
    \end{minipage}

    \vspace{2pt}
    \begin{minipage}{0.05\linewidth}
        \centering
        \rotatebox{90}{\textbf{scan65}}
    \end{minipage}
    \begin{minipage}{0.224\linewidth}
        \includegraphics[width=\linewidth]{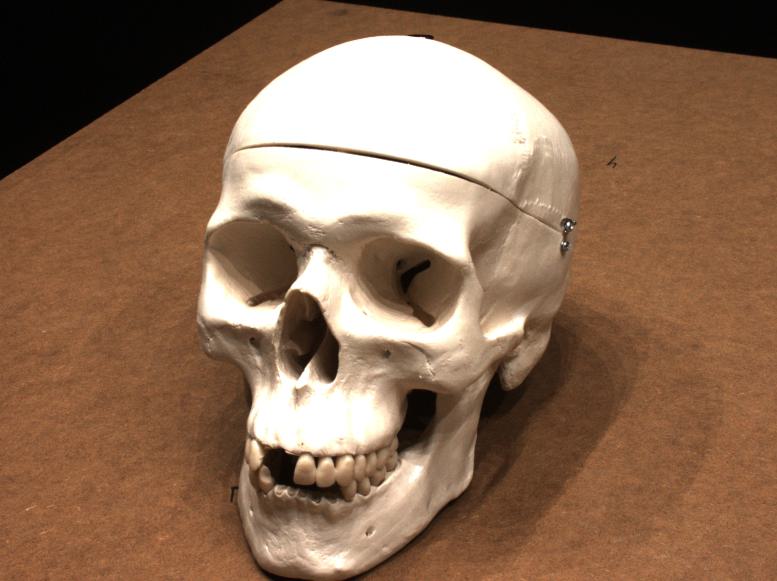}
    \end{minipage}
    \begin{minipage}{0.15652\linewidth}
        \begin{tikzpicture}
            \node[anchor=south west,inner sep=0] (image) at (0,0)
                {\includegraphics[width=\linewidth]{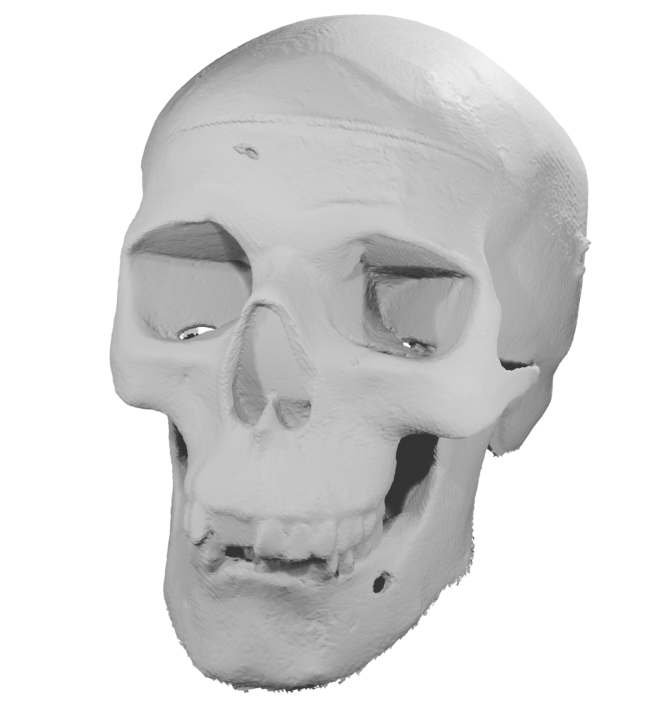}};
            
            \node[anchor=south west] at (1.44, 0.0) {\includegraphics[width=0.4\linewidth]{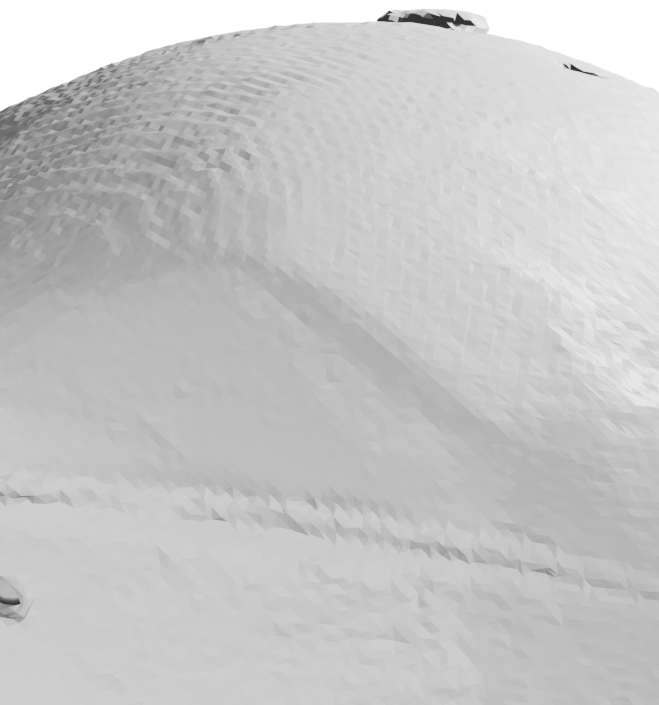}};
            \begin{scope}[x={(image.south east)},y={(image.north west)}]
                \draw[red, thick] (0.57,0.045) rectangle (0.97,0.44);
            \end{scope}
        \end{tikzpicture}
    \end{minipage}
    \begin{minipage}{0.15652\linewidth}
        \begin{tikzpicture}
            \node[anchor=south west,inner sep=0] (image) at (0,0)
                {\includegraphics[width=\linewidth]{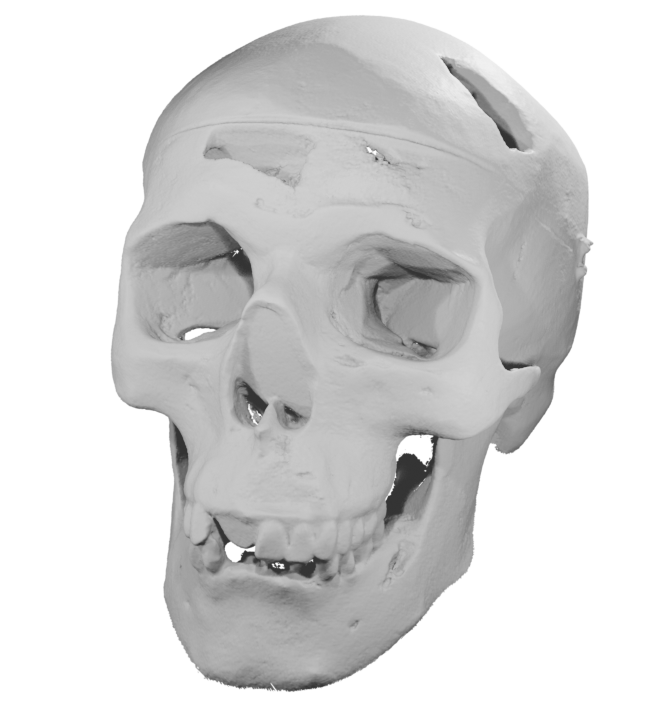}};
            
            \node[anchor=south west] at (1.44, 0.0) {\includegraphics[width=0.4\linewidth]{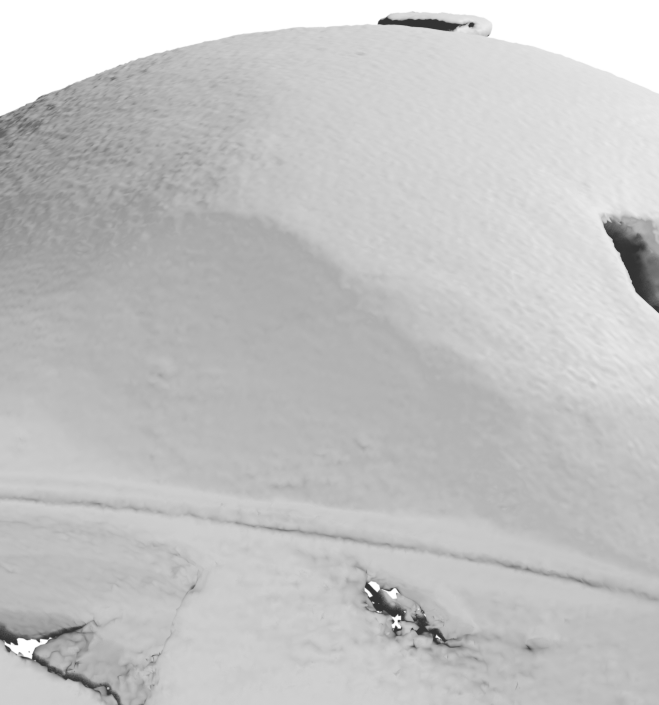}};
            \begin{scope}[x={(image.south east)},y={(image.north west)}]
                \draw[red, thick] (0.57,0.045) rectangle (0.97,0.44);
            \end{scope}
        \end{tikzpicture}
    \end{minipage}
    \begin{minipage}{0.15652\linewidth}
        \begin{tikzpicture}
            \node[anchor=south west,inner sep=0] (image) at (0,0)
                {\includegraphics[width=\linewidth]{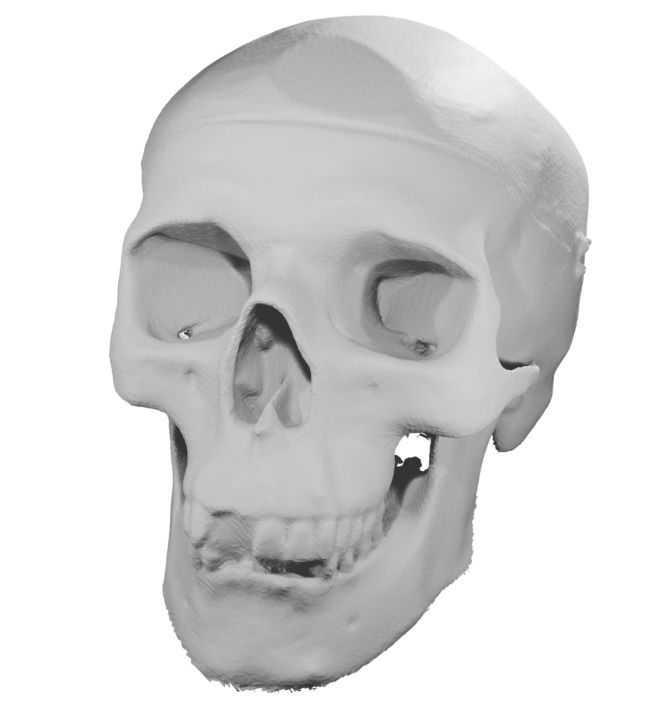}};
            
            \node[anchor=south west] at (1.44, 0.0) {\includegraphics[width=0.4\linewidth]{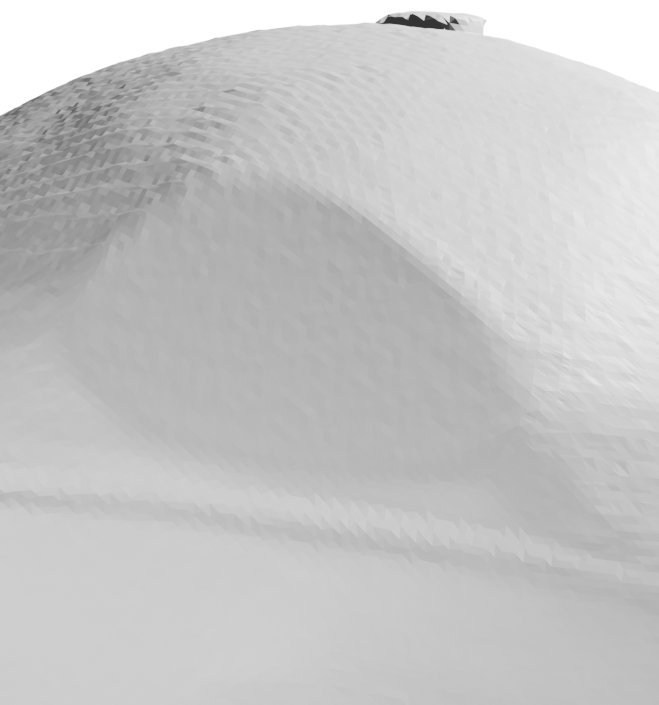}};
            \begin{scope}[x={(image.south east)},y={(image.north west)}]
                \draw[red, thick] (0.57,0.045) rectangle (0.97,0.44);
            \end{scope}
        \end{tikzpicture}
    \end{minipage}
    \begin{minipage}{0.15652\linewidth}
        \begin{tikzpicture}
            \node[anchor=south west,inner sep=0] (image) at (0,0)
                {\includegraphics[width=\linewidth]{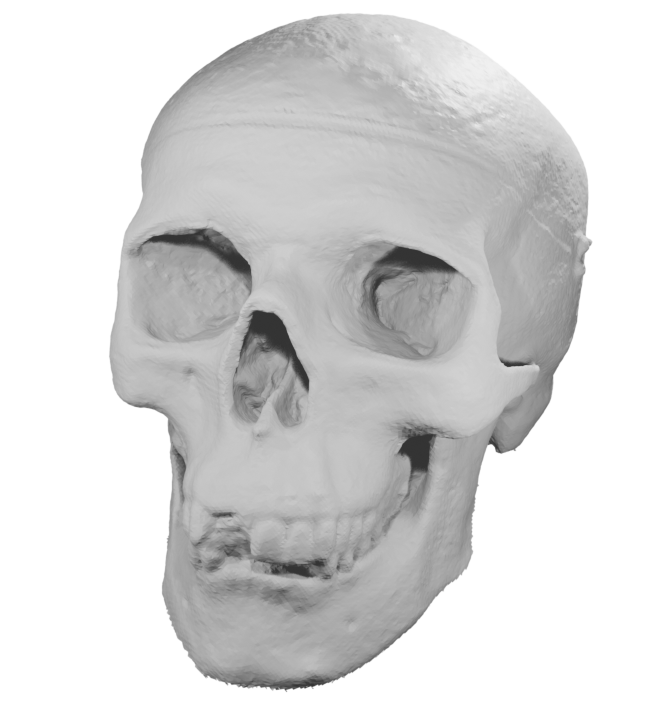}};
            
            \node[anchor=south west] at (1.44, 0.0) {\includegraphics[width=0.4\linewidth]{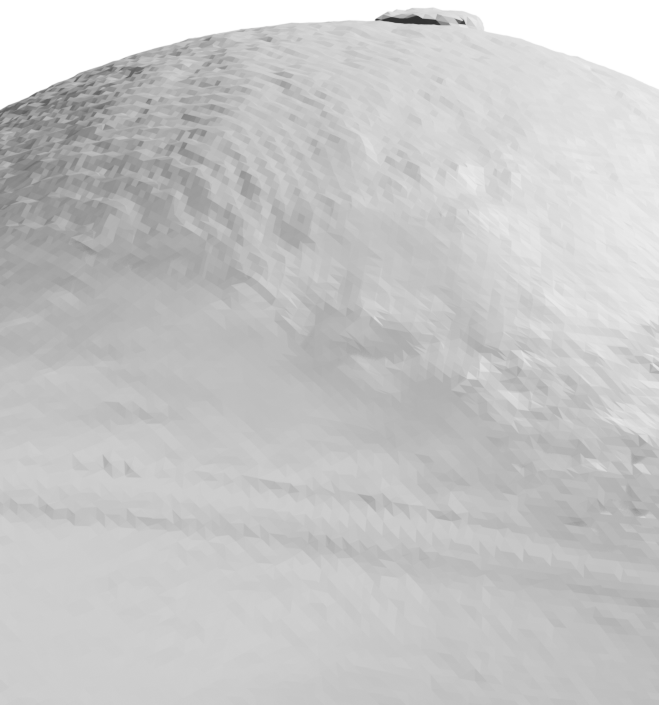}};
            \begin{scope}[x={(image.south east)},y={(image.north west)}]
                \draw[red, thick] (0.57,0.045) rectangle (0.97,0.44);
            \end{scope}
        \end{tikzpicture}
    \end{minipage}

    \vspace{2pt}
    \begin{minipage}{0.05\linewidth}
        \centering
        \rotatebox{90}{\textbf{scan118}}
    \end{minipage}
    \begin{minipage}{0.224\linewidth}
        \includegraphics[width=\linewidth]{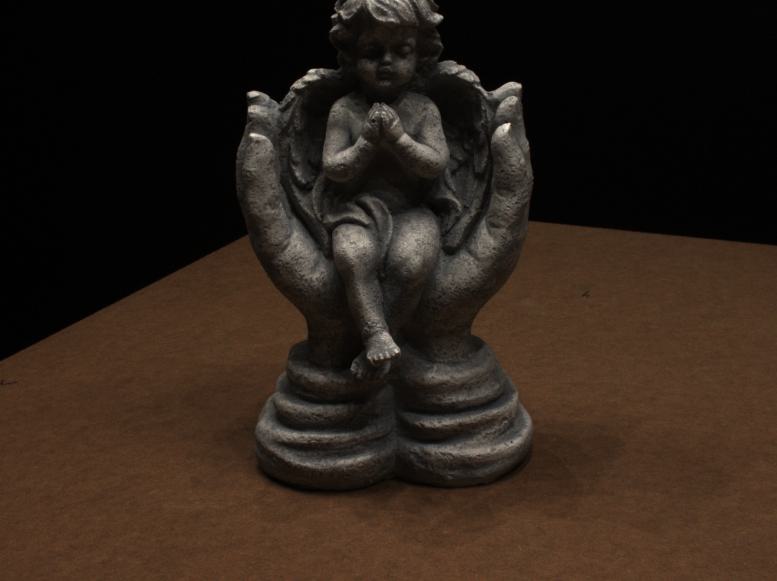}
        \vspace{-10pt}
        \caption*{Reference Image}
    \end{minipage}
    \begin{minipage}{0.15652\linewidth}
        \begin{tikzpicture}
            \node[anchor=south west,inner sep=0] (image) at (0,0)
                {\includegraphics[width=\linewidth]{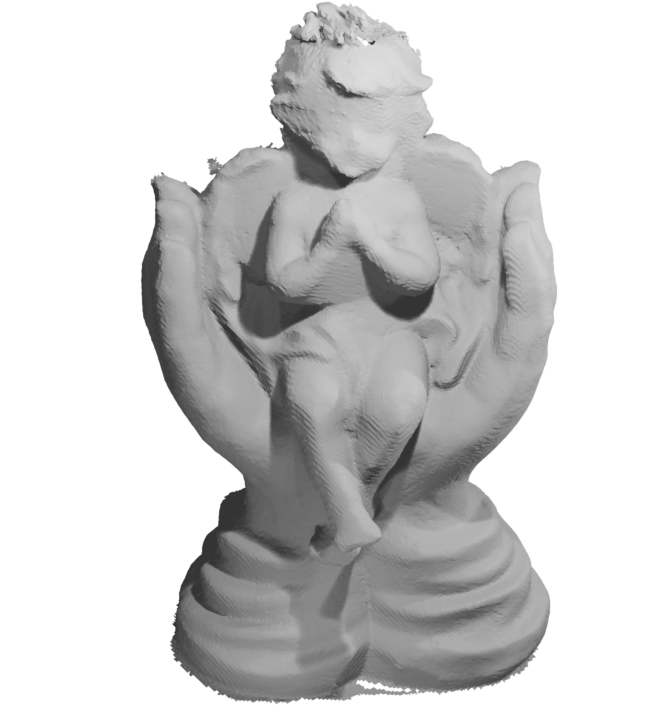}};
            
            \node[anchor=south west] at (1.44, 0.0) {\includegraphics[width=0.4\linewidth]{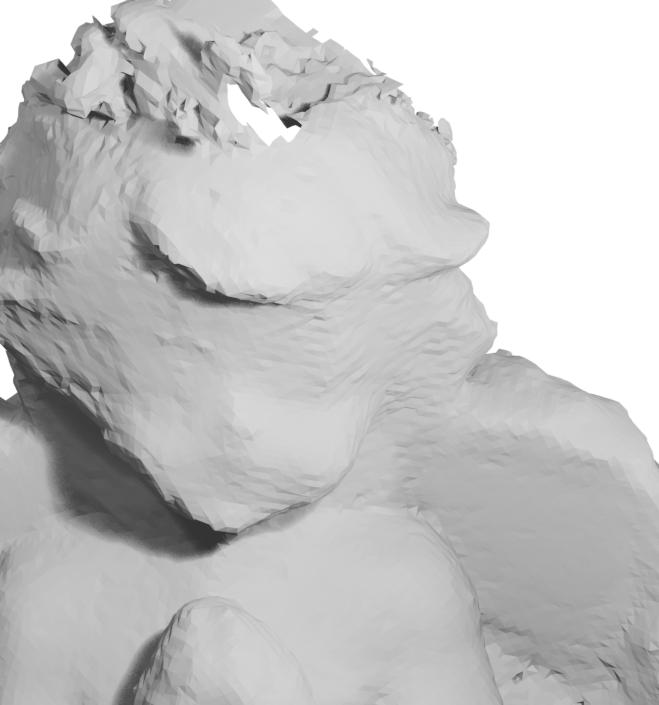}};
            \begin{scope}[x={(image.south east)},y={(image.north west)}]
                \draw[red, thick] (0.57,0.045) rectangle (0.97,0.44);
            \end{scope}
        \end{tikzpicture}
        \vspace{-10pt}
        \caption*{2DGS}
    \end{minipage}
    \begin{minipage}{0.15652\linewidth}
        \begin{tikzpicture}
            \node[anchor=south west,inner sep=0] (image) at (0,0)
                {\includegraphics[width=\linewidth]{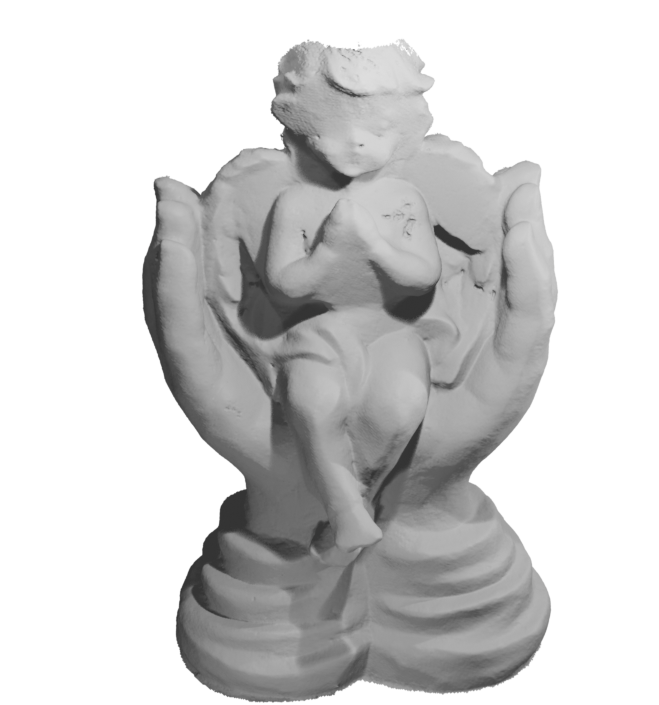}};
            
            \node[anchor=south west] at (1.44, 0.0) {\includegraphics[width=0.4\linewidth]{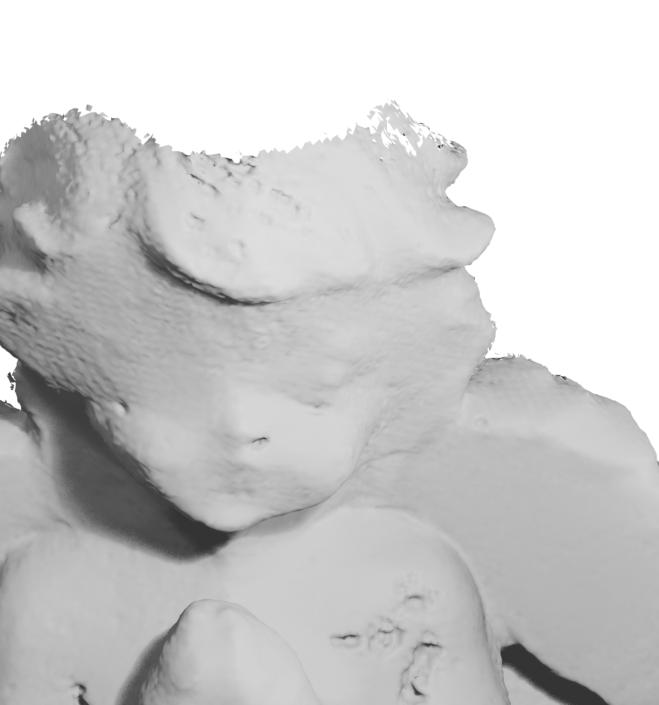}};
            \begin{scope}[x={(image.south east)},y={(image.north west)}]
                \draw[red, thick] (0.57,0.045) rectangle (0.97,0.44);
            \end{scope}
        \end{tikzpicture}
        \vspace{-10pt}
        \caption*{GOF}
    \end{minipage}
    \begin{minipage}{0.15652\linewidth}
        \begin{tikzpicture}
            \node[anchor=south west,inner sep=0] (image) at (0,0)
                {\includegraphics[width=\linewidth]{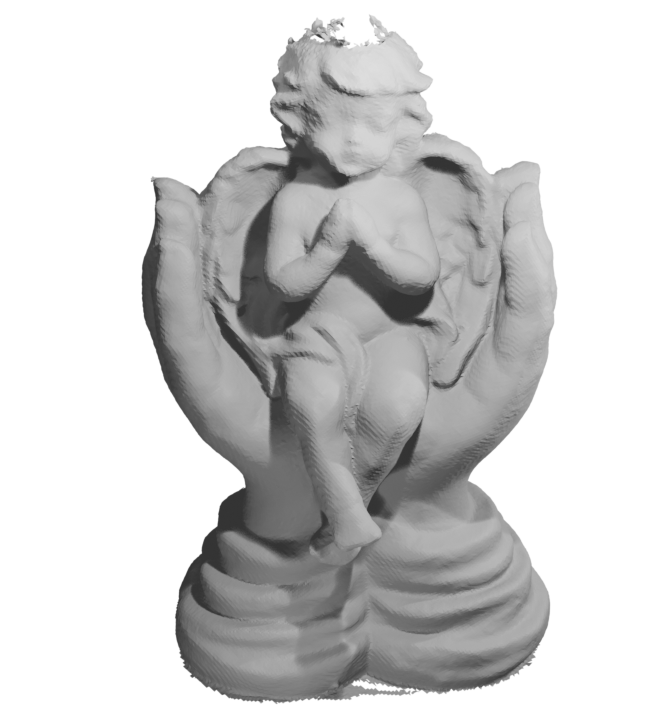}};
            
            \node[anchor=south west] at (1.44, 0.0) {\includegraphics[width=0.4\linewidth]{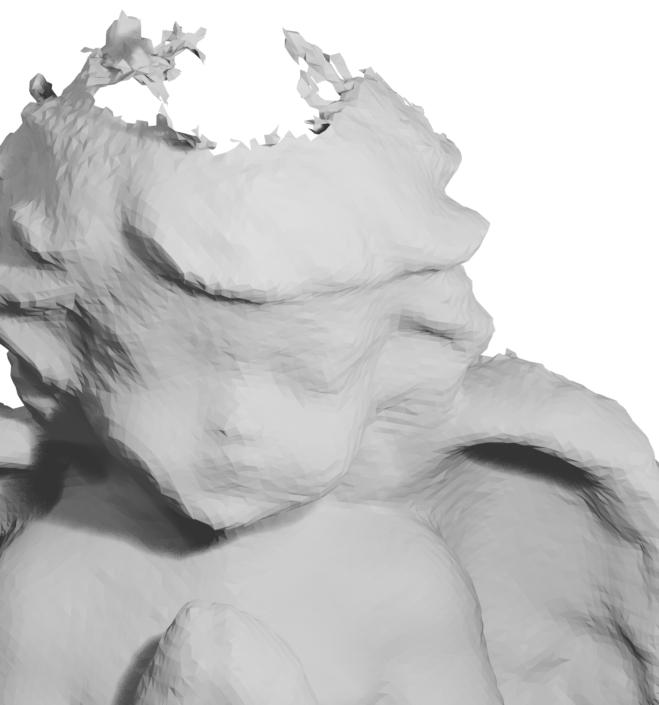}};
            \begin{scope}[x={(image.south east)},y={(image.north west)}]
                \draw[red, thick] (0.57,0.045) rectangle (0.97,0.44);
            \end{scope}
        \end{tikzpicture}
        \vspace{-10pt}
        \caption*{PGSR}
    \end{minipage}
    \begin{minipage}{0.15652\linewidth}
        \begin{tikzpicture}
            \node[anchor=south west,inner sep=0] (image) at (0,0)
                {\includegraphics[width=\linewidth]{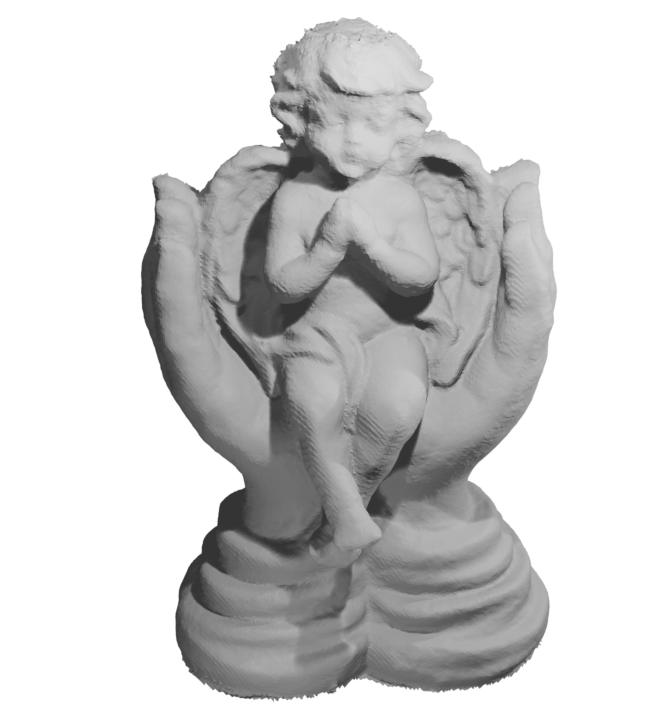}};
            
            \node[anchor=south west] at (1.44, 0.0) {\includegraphics[width=0.4\linewidth]{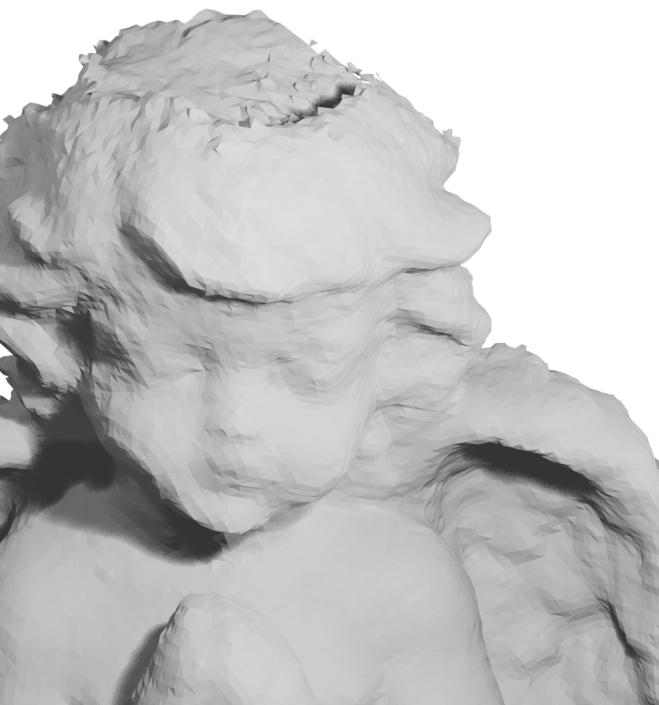}};
            \begin{scope}[x={(image.south east)},y={(image.north west)}]
                \draw[red, thick] (0.57,0.045) rectangle (0.97,0.44);
            \end{scope}
        \end{tikzpicture}
        \vspace{-10pt}
        \caption*{Ours}
    \end{minipage}

    \vspace{-7pt}
    \caption{Qualitative geometric reconstruction comparisons on the DTU dataset. Our method achieves reconstructions of higher quality and greater detail. }
    \label{fig:qualitative}
    \vspace{-5pt}
\end{figure*}

\begin{table*}
\centering
\renewcommand{\arraystretch}{0.9}
\resizebox{\textwidth}{!}{%
\begin{tabular}{l|ccccccccccccccc|cc}
\toprule
CD (mm)$ \downarrow$& 24 & 37 & 40 & 55 & 63 & 65 & 69 & 83 & 97 & 105 & 106 & 110 & 114 & 118 & 122 & Mean & Time\\
\midrule
NeuS \cite{wang2021neus} &1.00&1.37&0.93&0.43&1.10&0.65&0.57&1.48&1.09&0.83&0.52&1.20&0.35&0.49&0.54&0.84&$>$12h\\
VolSDF \cite{yariv2021volume} &1.14&1.26&0.81&0.49&1.25&0.70&0.72&1.29&1.18&0.70&0.66&1.08&0.42&0.61&0.55&0.86&$>$12h\\
Neuralangelo \cite{li2023neuralangelo} &\cellcolor{best!37}0.37&0.72&\cellcolor{best!37}0.35&\cellcolor{best!37}0.35&0.87&\cellcolor{best!37}0.54&\cellcolor{second!20}0.53&1.29&0.97&0.73&\cellcolor{second!20}0.47&\cellcolor{third!20}0.74&\cellcolor{best!37}0.32&\cellcolor{best!37}0.41&\cellcolor{third!20}0.43&\cellcolor{third!20}0.61&$>$128h\\
\midrule
3DGS \cite{kerbl3Dgaussians}&2.14&1.53&2.08&1.68&3.49&2.21&1.43&2.07&2.22&1.75&1.79&2.55&1.53&1.52&1.50&1.96&11.2min\\
Gaussian surfels~\cite{dai2024high}&0.66&0.93&0.54&0.41&1.06&1.14&0.85&1.29&1.53&0.79&0.82&1.58&0.45&0.66&0.53&0.88&\cellcolor{best!37}6.7min\\
SuGaR\cite{guedon2024sugar}&1.47&1.33&1.13&0.61&2.25&1.71&1.15&1.63&1.62&1.07&0.79&2.45&0.98&0.88&0.79&1.33&1h\\
2DGS \cite{huang20242d} &0.48&0.91&\cellcolor{third!20}0.39&0.39&1.01&0.83&0.81&1.36&1.27&0.76&0.70&1.40&0.40&0.76&0.52&0.80&\cellcolor{second!20}19.2min\\
GOF \cite{yu2024gaussian} &0.50&0.82&\cellcolor{second!20}0.37&\cellcolor{second!20}0.37&1.12&0.74&\cellcolor{third!20}0.73&\cellcolor{second!20}1.18&1.29&0.68&0.77&0.90&0.42&0.66&0.49&0.74&1h\\
GSDF \cite{yu2024gsdf}&0.59&0.94&0.46&\cellcolor{third!20}0.38&1.30&0.77&\cellcolor{third!20}0.73&1.59&1.29&0.76&0.59&1.22&0.38&\cellcolor{third!20}0.52&0.51&0.80&32min\\
GS-pull \cite{zhang2025neural}&0.51&\cellcolor{best!37}0.56&0.46&0.39&\cellcolor{third!20}0.82&0.67&0.85&1.37&1.25&0.73&0.54&1.39&\cellcolor{third!20}0.35&0.88&0.42&0.75&\cellcolor{third!20}22min\\
Ours w/o MV&0.46&0.76&0.40&\cellcolor{third!20}0.38&0.92&0.80&0.76&\cellcolor{third!20}1.25&\cellcolor{third!20}0.95&\cellcolor{third!20}0.67&0.62&1.20&0.38&0.60&0.47&0.71&25min\\
\midrule
PGSR \cite{chen2024pgsr}&\cellcolor{third!20}0.40&\cellcolor{second!20}0.60&\cellcolor{third!20}0.39&\cellcolor{second!20}0.37&\cellcolor{second!20}0.78&\cellcolor{third!20}0.59&\cellcolor{second!20}0.53&\cellcolor{second!20}1.18&\cellcolor{best!37}0.67&\cellcolor{second!20}0.63&\cellcolor{third!20}0.48&\cellcolor{second!20}0.62&\cellcolor{second!20}0.34&\cellcolor{second!20}0.42&\cellcolor{best!37}0.39&\cellcolor{second!20}0.56&40min\\
Ours&\cellcolor{second!20}0.38&\cellcolor{third!20}0.62&\cellcolor{second!20}0.37&\cellcolor{third!20}0.38&\cellcolor{best!37}0.75&\cellcolor{second!20}0.55&\cellcolor{best!37}0.51&\cellcolor{best!37}1.12&\cellcolor{second!20}0.68&\cellcolor{best!37}0.61&\cellcolor{best!37}0.46&\cellcolor{best!37}0.58&\cellcolor{third!20}0.35&\cellcolor{best!37}0.41&\cellcolor{second!20}0.40&\cellcolor{best!37}0.54&48min\\
\bottomrule
\end{tabular}
}
\vspace{-5pt}
\caption{Quantitative Chamfer Distance comparison on the DTU dataset. Our method achieves the best performance both with and without multi-view regularization. \raisebox{0.7ex}{\colorbox{best!37}{~}}, \raisebox{0.7ex}{\colorbox{second!20}{~}}, \raisebox{0.7ex}{\colorbox{third!20}{~}} denote the best, second best, and third best results, respectively.}
\label{exp:dtu}
\vspace{-10pt}
\end{table*}

\begin{table}
\centering
\resizebox{\columnwidth}{!}{%
\begin{tabular}{l|cc|ccc|cc}
\toprule
F1-Score $\uparrow$ & Geo-NeuS & N-angelo & 2DGS & GOF& Ours w/o MV &PGSR & Ours\\
\bottomrule
Barn &0.33&\cellcolor{best!37}0.70&0.41&0.51&0.46&\cellcolor{third!20}0.52&\cellcolor{second!20}0.55\\
Caterpillar &0.26&0.36&0.24&\cellcolor{best!37}0.41&0.32&\cellcolor{third!20}0.38&\cellcolor{second!20}0.40\\
Courthouse &0.12&\cellcolor{best!37}0.28&\cellcolor{third!20}0.16&\cellcolor{best!37}0.28&\cellcolor{second!20}0.26&\cellcolor{second!20}0.26&\cellcolor{best!37}0.28\\
Ignatius &0.72&\cellcolor{best!37}0.89&0.52&0.68&\cellcolor{third!20}0.79&0.77&\cellcolor{second!20}0.81\\
Meetingroom &0.20&\cellcolor{best!37}0.32&0.17&0.28&0.25&\cellcolor{third!20}0.29&\cellcolor{second!20}0.31\\
Truck &0.45&0.48&0.45&0.58&\cellcolor{third!20}0.60&\cellcolor{second!20}0.62&\cellcolor{best!37}0.64\\
Mean &0.35&\cellcolor{best!37}0.50&0.33&\cellcolor{third!20}0.46&0.45&\cellcolor{second!20}0.47&\cellcolor{best!37}0.50\\
Time &$>$24h&$>$127h&\cellcolor{best!37}34min&114min&\cellcolor{second!20}43min&\cellcolor{third!20}66min&75min\\

\bottomrule
\end{tabular}%
}
\vspace{-5pt}
\caption{Quantitative F1-Score comparison of QGS with GS-like and NeRF-like methods on the TNT dataset. QGS outperforms all methods, achieving state-of-the-art reconstruction results. }
\label{exp:tnt}
\vspace{-10pt}
\end{table}
We compared our method with several SOTA approaches across multiple datasets, including DTU \cite{jensen2014large}, TNT \cite{knapitsch2017tanks}, and Mip-NeRF 360 \cite{barron2022mip}. 
We evaluated geometry with F1-score and Chamfer Distance, and appearance with PSNR, SSIM, and LPIPS, followed by analysis and conclusions.
\subsection{Implementation Details. }
\noindent\textbf{Rasterizer.}
We implemented Quadratic Splatting with custom CUDA kernels on the 3DGS framework \cite{kerbl3Dgaussians}, extending the renderer to output depth distortion maps, depth maps, normal maps, and curvature maps. 
Since quadrics can be non-convex, we used rectangular truncation and approximations to compute image bounding boxes, as detailed in the supplementary materials. \par
\noindent\textbf{Settings. }
We adopted the adaptive control strategy from 3DGS \cite{kerbl3Dgaussians}. 
Similar to 2DGS \cite{huang20242d}, QGS projects the 3D center gradient onto screen space instead of using the 2D projected gradient. 
A gradient threshold of 0.3 and a percent dense value of 0.001 ensure consistent point cloud number with other methods.
All experiments were conducted on a single A6000 GPU.\par
\noindent\textbf{Mesh Extraction. }
We used median depth (i.e., $t_\mathrm{median} = \max{t_i | T_i > 0.5}$) as the final output depth.
Then we fused the depth maps using Truncated Signed Distance Fusion (TSDF) with Open3D \cite{zhou2018open3d}. 
Following the 2DGS setup, we set the voxel size to 0.004 and truncation threshold to 0.02.\par
\subsection{Comparison.}
\noindent\textbf{Geometry Evaluation.}
In all experiments, evaluations were performed at half the image resolution. 
To clearly demonstrate the improvement introduced by our quadric surfels and to maintain consistency with other methods, we denote the variant of QGS without multi-view regularization loss as QGS w/o MV. 
In Table \ref{exp:dtu}, we compared QGS with implicit \cite{wang2021neus,yariv2021volume,li2023neuralangelo} and explicit \cite{huang20242d,yu2024gaussian,guedon2024sugar,yu2024gsdf,chen2024pgsr,zhang2025neural} SOTA reconstruction methods on the DTU dataset using Chamfer Distance as the metric. 
As shown in Table \ref{exp:dtu}, our method outperforms both implicit and explicit approaches on the DTU dataset. 
Furthermore, even without multi-view regularization, our method surpasses others in accuracy while maintaining competitive speed. 
This improvement stems from the superior geometric fitting capability of quadrics, which enables finer detail preservation, as illustrated in Fig~\ref{fig:qualitative}.

\begin{figure}[!t]
    \centering
    \begin{minipage}{0.49\linewidth}     
        \includegraphics[width=1\linewidth]{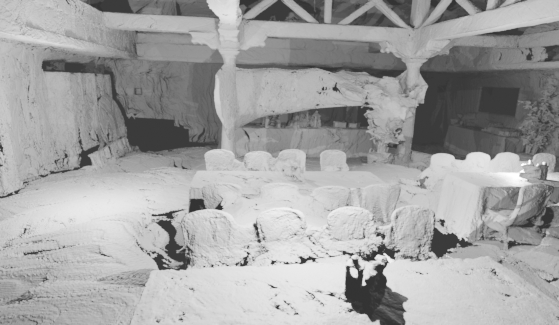}
        \includegraphics[width=1\linewidth]{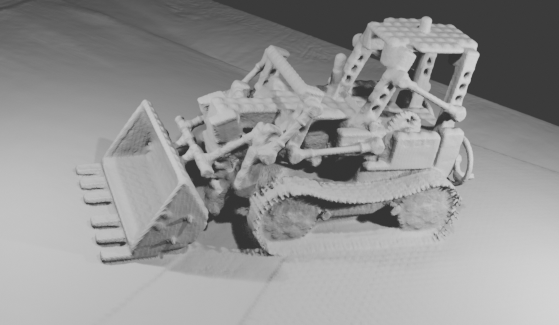}
        \vspace{-15pt}
        \caption*{(a) 2DGS}
    \end{minipage}
    \begin{minipage}{0.49\linewidth}
        \includegraphics[width=1\linewidth]{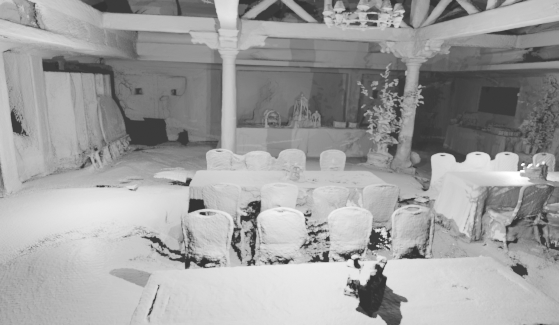}
        \includegraphics[width=1\linewidth]{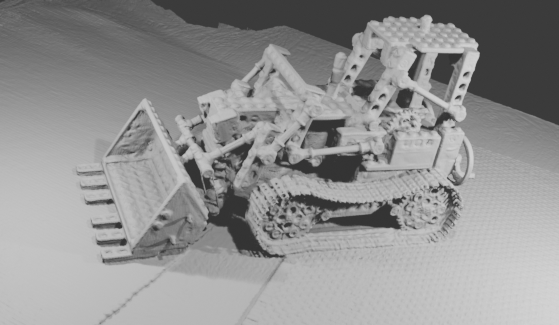}
        \vspace{-15pt}
        \caption*{(b) QGS w/o MV}
    \end{minipage}\vspace{-6pt}
    \caption{Mesh reconstruction comparison between 2DGS and QGS w/o MV. The first row shows the Meetingroom in TNT dataset, and the second row shows the kitchen scene in Mip-NeRF 360 dataset. }
    \label{fig:2DGS_QGS}
\end{figure}

In Table \ref{exp:tnt}, we also evaluated QGS and other SOTA methods \cite{fu2022geo,li2023neuralangelo,huang20242d,yu2024gaussian,chen2024pgsr} on the TNT dataset using F1-score.
Without multi-view regularization, our method significantly outperforms the disk-based 2DGS and achieves comparable results to GOF while requiring only half the computation time. 
With multi-view regularization, it surpasses all methods in reconstruction accuracy. 
Additionally, we provide a qualitative comparison between QGS without multi-view regularization and 2DGS to further demonstrate the enhanced geometric fitting capability of quadric surfels, as shown in Fig \ref{fig:2DGS_QGS}.\par
\par

\noindent\textbf{Rendering Evaluation.}
We compared rendering quality on the Mip-NeRF360 dataset with baseline approaches.  
As a surface-based method, QGS achieves competitive rendering results in indoor scenes, as shown in Table \ref{exp:appearance}, but performs less effectively in outdoor environments. 
We attribute this to QGS’s higher geometric fitting capability, which may lead to overfitting in regions with sparse views or low texture. 
Future work could explore additional regularization constraints, particularly on the curvature of quadric surfaces, to address this issue.\par
\begin{table}
\centering
\renewcommand{\arraystretch}{0.9}
\resizebox{\columnwidth}{!}{\begin{tabular}{l|ccc|ccc}
\toprule
& \multicolumn{3}{c}{Indoor scenes} & \multicolumn{3}{c}{Outdoor scenes} \\
&PSNR$\uparrow$ & SSIM$\uparrow$ & LPIPS$\downarrow$ & PSNR$\uparrow$ & SSIM$\uparrow$ & LPIPS$\downarrow$\\
\midrule
NeRF&26.84&0.790&0.370&21.46&0.458&0.515\\
Deep Blending&26.40&0.844&0.261&21.54&0.524&0.364\\
i-NGP&29.15&0.880&0.216&22.90&0.566&0.371\\
Mip-NeRF360&\cellcolor{best!37}31.72&0.917&0.180&\cellcolor{third!20}24.47&0.691&0.283\\
\midrule
3DGS&\cellcolor{third!20}30.52&0.921&0.199&24.45&\cellcolor{second!20}0.728&0.240\\
SuGar&29.44&0.911&0.216&22.76&0.631&0.349\\
2DGS&30.39&\cellcolor{third!20}0.924&0.182&24.33&0.709&0.284\\
GOF&\cellcolor{second!20}30.80&\cellcolor{best!37}0.928&\cellcolor{second!20}0.167&\cellcolor{best!37}24.76&\cellcolor{best!37}0.742&\cellcolor{best!37}0.225\\
Ours w/o MV&30.48&\cellcolor{second!20}0.926&\cellcolor{best!37}0.166&\cellcolor{second!20}24.56&\cellcolor{third!20}0.724&\cellcolor{third!20}0.239\\
\midrule
PGSR&30.35&\cellcolor{third!20}0.924&\cellcolor{third!20}0.176&24.29&0.718&\cellcolor{second!20}0.236\\
Ours&30.45&0.919&0.184&24.32&0.706&0.242\\
\bottomrule
\end{tabular}}
\caption{Quantitative comparison of appearance between QGS, GS-like, and NeRF-like methods on the Mip-NeRF 360 dataset.}
\label{exp:appearance}
\vspace{-2pt}
\end{table}
\subsection{Ablation}
\begin{figure}
    \centering
    \begin{minipage}{0.24\linewidth}
        \includegraphics[width=1\linewidth]{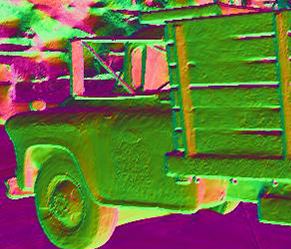}
        \vspace{-15pt}
        \caption*{(a) w/o $\lambda_K$}
    \end{minipage}
    \begin{minipage}{0.24\linewidth}
        \includegraphics[width=1\linewidth]{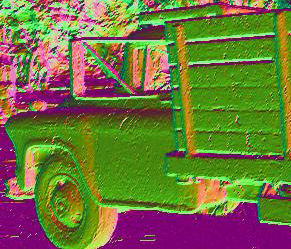}
        \vspace{-15pt}
        \caption*{(b) w/o Sort}
    \end{minipage}
    \begin{minipage}{0.24\linewidth}
        \includegraphics[width=1\linewidth]{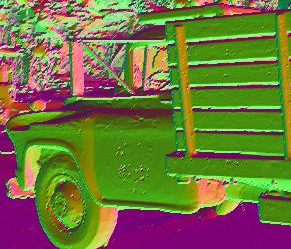}
        \vspace{-15pt}
        \caption*{(c) w/o MV}
    \end{minipage}
    \begin{minipage}{0.24\linewidth}
        \includegraphics[width=1\linewidth]{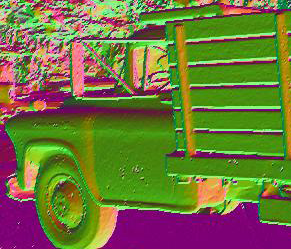}
        \vspace{-15pt}
        \caption*{(d) Full}
    \end{minipage}
    \vspace{-5pt}
    \caption{Ablation study. The full setting achieves the highest reconstruction quality.}
    \label{fig:ablation}
    \vspace{-2pt}
\end{figure}

In this section, we assess the impact of individual QGS components on reconstruction quality, including per-pixel resorting, curvature-guided normal consistency, and multi-view regularization, as presented in Table \ref{exp:ablation_wMV} and Fig \ref{fig:ablation}. 
We denote w/o sort as the absence of per-pixel resorting, w/o $\lambda_K$ as the absence of curvature-guided normal consistency, and w/o MV as the absence of multi-view regularization.
We observe that: (a) Disabling curvature guidance fills fine gaps in the normal map, reducing reconstruction quality. 
(b) The absence of per-pixel sorting similarly fills gaps and leads to uneven surfaces.
(c) Disabling multi-view regularization causes overfitting in low-texture regions, resulting in surface artifacts, such as dents on the front of the truck. \par
Additionally, we conducted ablation studies without multi-view regularization to isolate and evaluate the effectiveness of quadrics.  
Specifically, we fixed the quadric's z-axis scale to 0.001, approximating it as a disk (denoted as $s_3$ fixed). 
We also evaluated optimization using Euclidean distance (denoted as w/ Euclid). 
Both modifications significantly reduced reconstruction quality, as shown in Table \ref{exp:ablation}.
\par
\begin{table}[!t]
\centering
\renewcommand{\arraystretch}{0.9}
\resizebox{\columnwidth}{!}{
\begin{tabular}{l|ccccccc}
\toprule
F1-Score $\uparrow$&B&C&CH&I&M&T&mean \\
\midrule
Full model&\cellcolor{second!20}0.55&\cellcolor{best!37}0.40&\cellcolor{best!37}0.28&\cellcolor{best!37}0.81&\cellcolor{best!37}0.31&\cellcolor{best!37}0.64&\cellcolor{best!37}0.50\\
w/o Sort&\cellcolor{third!20}0.53&\cellcolor{second!20}0.39&\cellcolor{third!20}0.26&\cellcolor{third!20}0.77&\cellcolor{second!20}0.29&\cellcolor{second!20}0.62&\cellcolor{third!20}0.48\\
w/o $\lambda_K$&\cellcolor{best!37}0.56&\cellcolor{third!20}0.37&\cellcolor{second!20}0.27&\cellcolor{second!20}0.80&\cellcolor{second!20}0.29&\cellcolor{second!20}0.62&\cellcolor{second!20}0.49\\
w/o MV&0.46&0.32&\cellcolor{third!20}0.26&0.79&\cellcolor{third!20}0.25&\cellcolor{third!20}0.60&0.45\\
\bottomrule
\end{tabular}
}
\caption{Quantitative ablation study on the TNT dataset. We present ablation results for Barn (B), Caterpillar (C), Courthouse (CH), Ignatius (I), Meeting Room (M), and Truck (T). Our full setting achieves the best performance.}
\label{exp:ablation_wMV}
\end{table}
\begin{table}[!t]
\centering
\small
\renewcommand{\arraystretch}{0.7}
\begin{tabular}{c|cccc}
\toprule
Setting&w/o MV&$s_3$ fixed&w/ Euclid\\
\midrule
Mean F1$\uparrow$&\textbf{0.45}&0.37&0.38\\
\bottomrule
\end{tabular}
\caption{Ablation study without multi-view regularization on the TNT dataset. Fixing the quadric as a disk or using Euclidean distance optimization both result in reduced reconstruction quality.}
\label{exp:ablation}
\vspace{-10pt}
\end{table}

%% file: sec/5_conclusion.tex
\vspace{-5pt}
\section{Conclusion}
\vspace{-5pt}
In this work, we introduce Quadratic Gaussian Splatting, a variant of GS-like methods, designed to reconstruct accurate scene geometry and recover finer details.
QGS is the first to introduce quadric surfaces to Gaussian Splatting, defining Gaussian distributions in non-Euclidean space to improve fitting and capture second-order curvature. 
We achieve SOTA geometric reconstruction and competitive rendering results on various indoor and outdoor datasets. 
Additional results are in the supplementary materials.\par
\vspace{-10pt}
\section*{Acknowledgement}
\vspace{-5pt}
This work was supported by the National Natural Science Foundation of China (Grants U22B2055 and 62273345), the Beijing Natural Science Foundation (Grant L223003), and the Key R\&D Project in Henan Province (Grant 231111210300).

%% file: sec/X_suppl.tex
\clearpage
\setcounter{page}{1}
\maketitlesupplementary

In this supplementary material, we provide detailed explanations of the following:
(1) Comparison of quadric (QGS) and disk (2DGS~\cite{huang20242d}). 
(2) Calculation of the projected bounding box of primitives.
(3) Details of per-tile and per-pixel sorting.
(4) Solving the ray-primitive intersection equation and numerical handling.
(5) Derivation of the geodesic distance formula integration.
(6) Derivation of Gaussian curvature.
(7) We present more qualitative results. 
\vspace{-10pt}
\vspace{-5pt}
\subsection*{Comparison of Disk and Quadric}
\begin{figure}[b]
    \vspace{-15pt}
    \centering
    \begin{minipage}{0.28\linewidth}
        \includegraphics[width=1\linewidth]{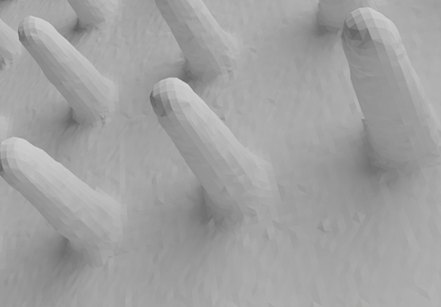}
    \vspace{-20pt}
    \caption*{(a) Mesh}
    \end{minipage}
    \begin{minipage}{0.28\linewidth}
        \includegraphics[width=1\linewidth]{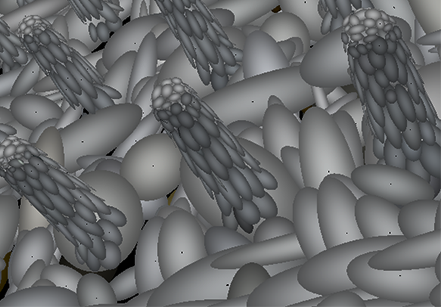}
    \vspace{-20pt}
    \caption*{(b) Disk}
    \end{minipage}
    \begin{minipage}{0.28\linewidth}
        \includegraphics[width=1\linewidth]{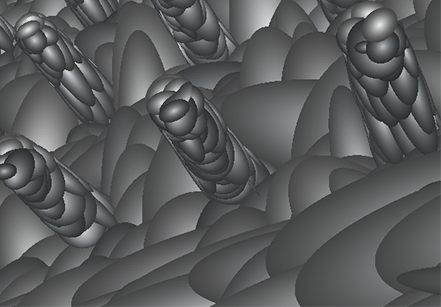}
    \vspace{-20pt}
    \caption*{(c) Quadric}
    \end{minipage}
    \vspace{-10pt}
    \caption{Visualizaion results on GSO dataset~\cite{downs2022google}. (a) shows the ground truth mesh, (b) displays the disk primitives, and (c) presents the quadric primitives.}
    \label{supp:GSO_visual}
    \vspace{-10pt}
\end{figure}

\vspace{-3pt}
We visualize the primitives of QGS (Quadric) and 2DGS (Disk), as shown in Fig~\ref{supp:GSO_visual}. 
It can be observed that quadrics tightly conform to the surface in a curved manner, with a more noticeable fit in regions of high curvature.
\vspace{-5pt}

\subsection*{Details of Bounding Box Calculation.}
\begin{figure}[!hb]
    \centering
    \includegraphics[width=1\linewidth]{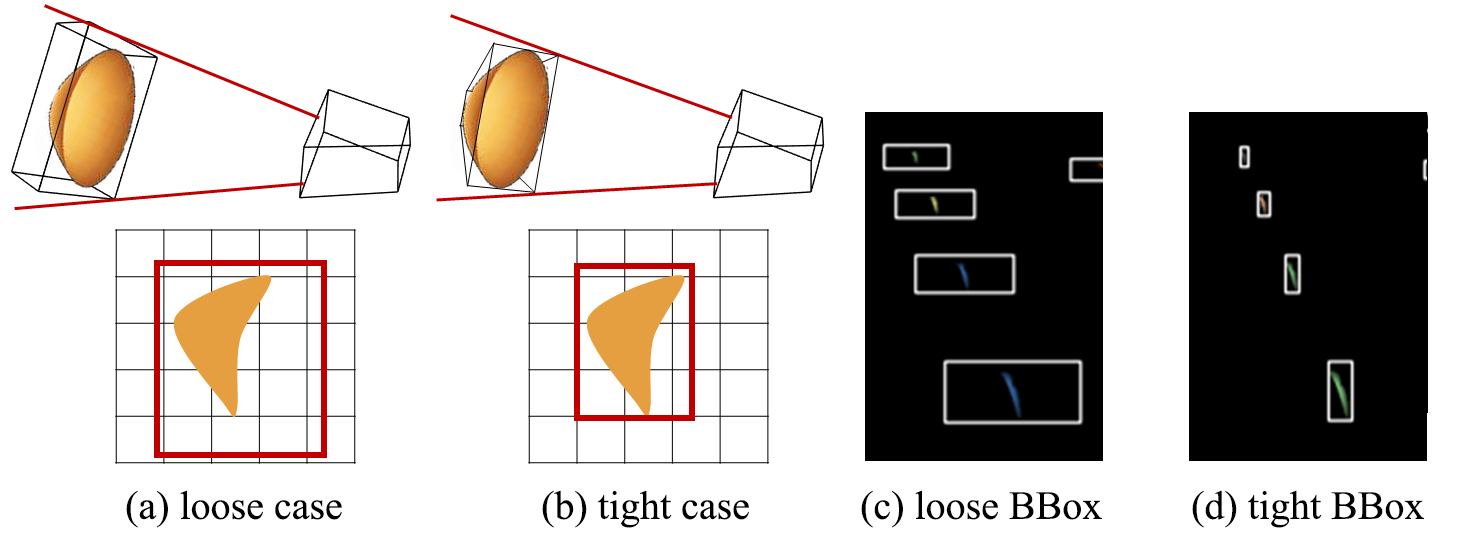}
    \caption{Comparison diagram of the bounding boxes between loose and tight case.}
    \vspace{-10pt}
    \label{fig:supp:bbox}
\end{figure}

\vspace{-3pt}
In 2DGS/3DGS \cite{huang20242d,kerbl3Dgaussians}, Gaussian primitives are enclosed convex representations, making it straightforward and convenient to compute their bounding box on the image, as shown in the first column of Fig \ref{fig:supp:bbox}.\par
\begin{figure}
    \centering
    \begin{minipage}{0.3\linewidth}
        \includegraphics[width=1\linewidth]{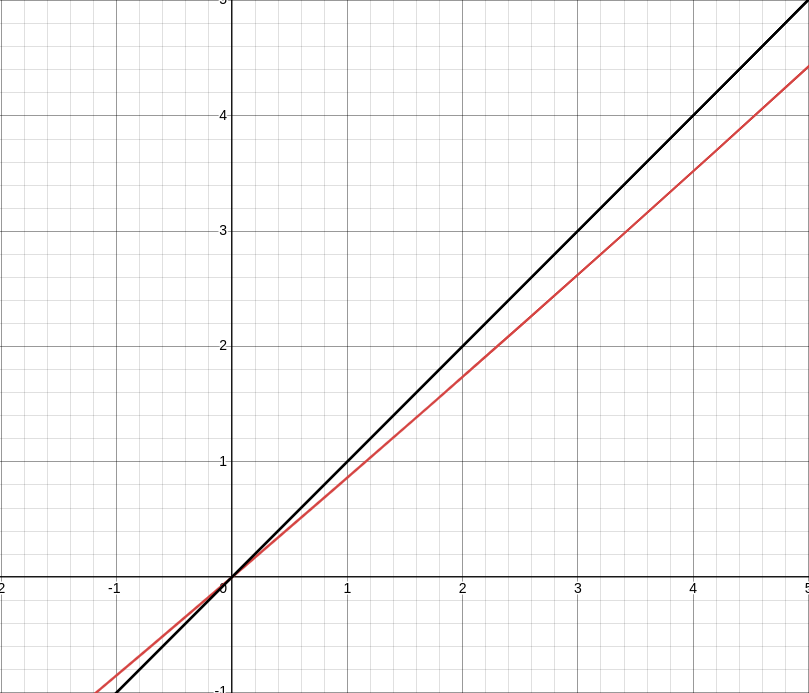}
        \vspace{-20pt}
        \caption*{$a=0.01$}
    \end{minipage}
    \begin{minipage}{0.3\linewidth}
        \includegraphics[width=1\linewidth]{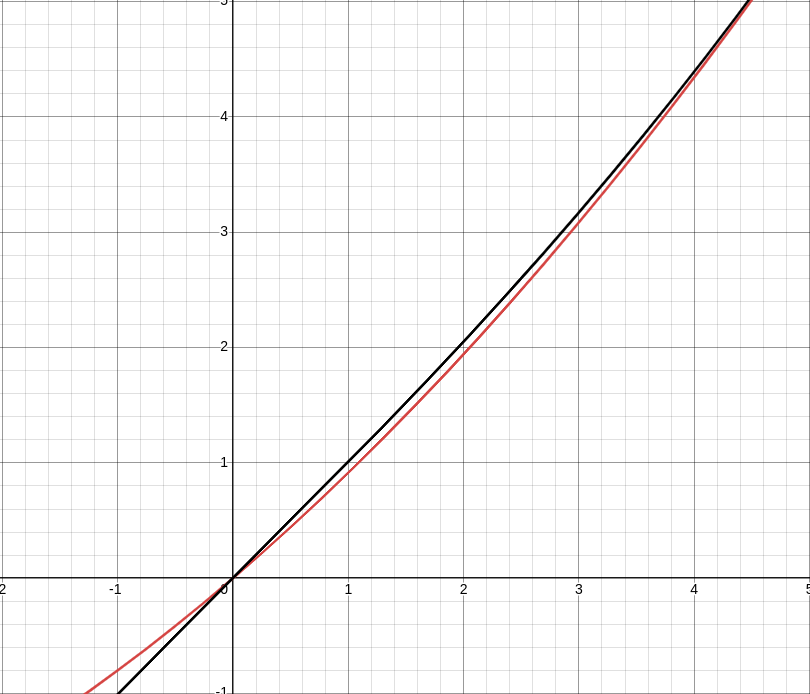}
        \vspace{-20pt}
        \caption*{$a=0.1$}
    \end{minipage}
    \begin{minipage}{0.3\linewidth}
        \includegraphics[width=1\linewidth]{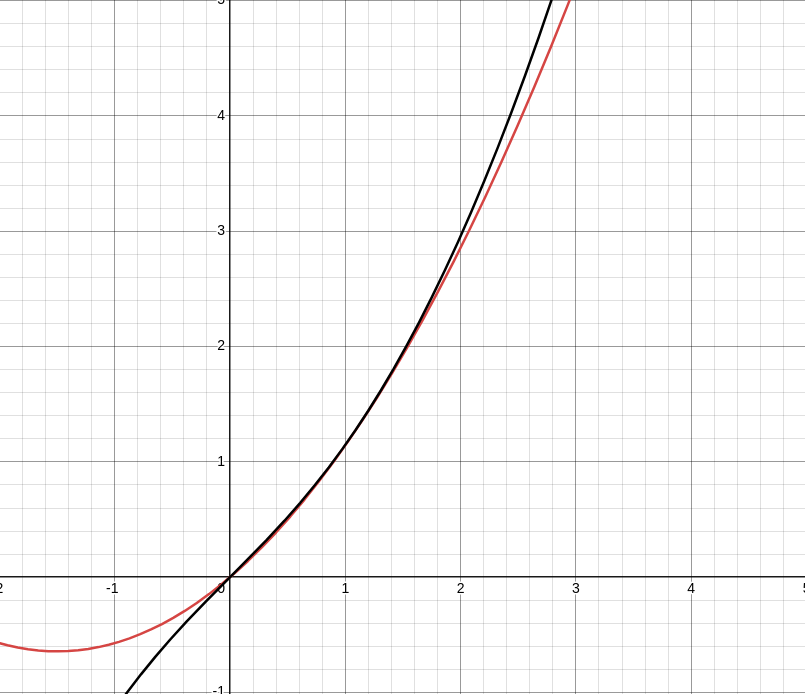}
        \vspace{-20pt}
        \caption*{$a=0.5$}
    \end{minipage}
    \vspace{-3pt}
    \caption{Comparison diagram of the approximate function versus the geodesic function. The red line represents the approximate function, while the black line represents the geodesic function.}
    \label{fig:supp:approximate_line}
    \vspace{-2pt}
\end{figure}

However, QGS includes concave cases, and geodesic distance complicates analytical determination of the Gaussian primitive's rendered portion during preprocessing. 
Specifically, the geodesic function Equation 10 in the main text lacks an elementary inverse, so computing the equipotential $\rho(\theta_0) = \{l^{-1}(l_0): l_0 = \sigma(\theta_0)\}$, requires numerical or approximate solutions. 
Moreover, even with an analytical solution for ${\rho(\theta_0) = l^{-1}(\sigma(\theta_0))}$, the Gaussian distribution in non-Euclidean space means equipotential lines may not fully enclose the primitive, requiring extra boundary calculations.
A straightforward approach is to use the projection of the quadric's 3D bounding box as the 2D bounding box, as shown in Fig \ref{fig:supp:bbox} (a). 
Its eight vertices are projected onto the image plane, and their 2D axis-aligned bounding box (AABB) is used as the 2D bounding box, as illustrated in Fig \ref{fig:supp:bbox} (c).
However, this loose case significantly enlarges the 2D bounding box, leading to a decrease in efficiency. 
To address this, we construct a tighter bounding box using the quadric’s tangent planes, as shown in Fig \ref{fig:supp:bbox} (b). 
Specifically, we first scale the geodesic function to a quadratic polynomial function:
\begin{equation}
    l(a,\rho)\approx\hat{l}(a,\rho)=\frac{4|a|\rho^2+6\rho}{7}
    \label{eq:approx}
\end{equation}
Given the direction $\theta_0$, we obtain the variance of the Gaussian distribution $\sigma_0(\theta_0)$. The root of the equation $\hat{l}(a,\rho) - \sigma_0 = 0$ in this direction can be calculated as:\par
\begin{equation}
    \rho_{0,1}=\frac{-6\pm\sqrt{36+112|a|\sigma_0}}{8|a|}
\end{equation}
The comparison between the approximate function and the geodesic distance function is shown in Fig \ref{fig:supp:approximate_line}, demonstrating that the two functions nearly overlap.\par
Since the Equation \ref{eq:approx} through the origin, we take its positive root, representing the horizontal distance from the primitive's vertex at a geodesic distance of $\sigma_0(\theta_0)$. 
We compute this for the major and minor axes, i.e. $\theta = 0$ and $\theta = \pi/2$, yielding $\rho_1$ and $\rho_2$, and obtain four points: $(\pm\rho_1,0,\rho_1^2),(0,\pm\rho_2,\rho_2^2)$. 
We then compute the intersection of the tangent planes at these four points with the plane $z=0$. 
\begin{equation}
\begin{aligned}
    x_1=\pm\frac{\rho_1s_1-\rho_1^2}{s_1},\;\;
    x_2=\pm\frac{\rho_2s_2-\rho_2^2}{s_2}  
\end{aligned}
\end{equation}
At this stage, we obtain a 3D truncated pyramid with top dimensions $\rho_1\times\rho_2$, bottom dimensions $x_1\times x_2$, and height $\max(\rho_1^2,\rho_2^2)$. 
We then project its eight vertices onto the image plane and determine the minimum 2D bounding box enclosing the resulting polygon as shown in Fig \ref{fig:supp:bbox} (d). \par
\begin{table}[!t]\scriptsize
\centering
\begin{tabular}{|c|c|c|c|c|c|}
\hline
& \multicolumn{3}{c|}{Mip-NeRF360} & \multicolumn{2}{c|}{TNT} \\
\hline
&Storage & FPS & Training Time & Storage & FPS \\
\hline
2DGS & 556MB & 15.34 & 1h5min & 218MB & 31.66\\
GOF & 825MB & 7.61 & 2h12min & 367MB & 13.45\\
QGS & 688MB & 7.92& 1h48min & 274MB & 13.27\\
QGS w/ TB & 641MB & 14.15& 1h13min & 263MB & 25.36\\
\hline
\end{tabular}
\caption{Speed and storage comparison on the Mip-NeRF 360~\cite{barron2022mip} and TNT~\cite{knapitsch2017tanks} datasets. With a tighter bounding box, QGS achieves a noticeable speed improvement.}
\label{tab:supp:speed_and_storage}
\vspace{-15pt}
\end{table}\par
By introducing a tighter bounding box and optimizing global memory access on the GPU, QGS achieves twice the speed. 
We denote the accelerated QGS as QGS w/ TB and compare the storage and speed of different methods on the Mip-NeRF 360 and TNT datasets as shown in Table \ref{tab:supp:speed_and_storage}.
\vspace{-5pt}
\subsection*{Details of Per-tile Sorting and Per-pixel Resorting}
\vspace{-3pt}
2DGS \cite{huang20242d} suggests the surface lies at the median intersection point where opacity reaches 0.5. 
However, for surface-based representations like 2DGS and QGS, the Gaussian distribution is concentrated on the surface, making the alpha-blending order more sensitive.\par
To address this, we introduce StopThePop's per-tile sorting and per-pixel resorting \cite{radl2024stopthepop} for more precise ordering. 
Specifically, for each $16 \times 16$ pixel tile, we compute the intersection depth using the ray from the pixel closest to the projected vertex of the quadric surface and apply this depth to all 256 pixels for tile-based global sorting. 
As shown in Fig 7 in the main text, this method effectively removes streak-like inconsistencies but introduces minor blocky artifacts due to approximating with a single ray per tile. 
Then we adopt StopThePop's per-pixel local resorting to reorder the Gaussians along each ray, to eliminate blocky inconsistencies. 
Specifically, after calculating each Gaussian's depth, normal, and other properties, we do not use them for alpha-blending immediately. 
Instead, we store them in an 8-length buffer array, and once the buffer is full, we select the closest Gaussian for alpha-blending. 
In original 3DGS, forward and backward computations use different traversal orders, causing the buffering mechanism to produce inconsistent rendering. 
Thus, following StopThePop's strategy, we perform gradient computation in a near-to-far order, which requires modifying the original gradient formulas. 
For volumetric rendering maps, we can uniformly express them as:
\begin{equation*}
    \hat{X}=\sum_{i=0}^{N-1}\bar{\alpha}_iT_iX_i,\;\;\mathrm{where}\;\bar{\alpha}_i=\alpha_ig_i,T_i=\prod_{j=0}^{i-1}(1-\bar{\alpha}_j)
\end{equation*}
Where $X_i$ represents properties such as the color, depth, normal, or curvature of the primitives. 
The gradient computation in a back-to-front order is given as follows:
\begin{equation*}
    \begin{aligned}
\frac{\partial{\hat{X}}}{\partial{\bar{\alpha}_{i}}}&=(X_i-\frac{\sum_{j=i+1}^N{X_j\bar{\alpha}_jT_j}}{T_{i+1}})T_i
\end{aligned}
\end{equation*}
We rewrite it as front-to-back form:
\begin{equation*}
    \begin{aligned}
\frac{\partial{\hat{X}}}{\partial{\bar{\alpha}_{i}}}=(X_i-\frac{\hat{X}-\sum_{j=0}^{i}{X_j\bar{\alpha}_jT_j}}{T_{i+1}})T_i
\end{aligned}
\end{equation*}
Ensuring that the backward computation follows the same order as the forward computation. 
Fortunately, the gradient computation order does not affect the distortion loss, meaning no additional derivation is needed for the distortion loss.
\vspace{-5pt}
\begin{figure}[!hb]
\vspace{-10pt}
    \centering
    \includegraphics[width=0.75\linewidth]{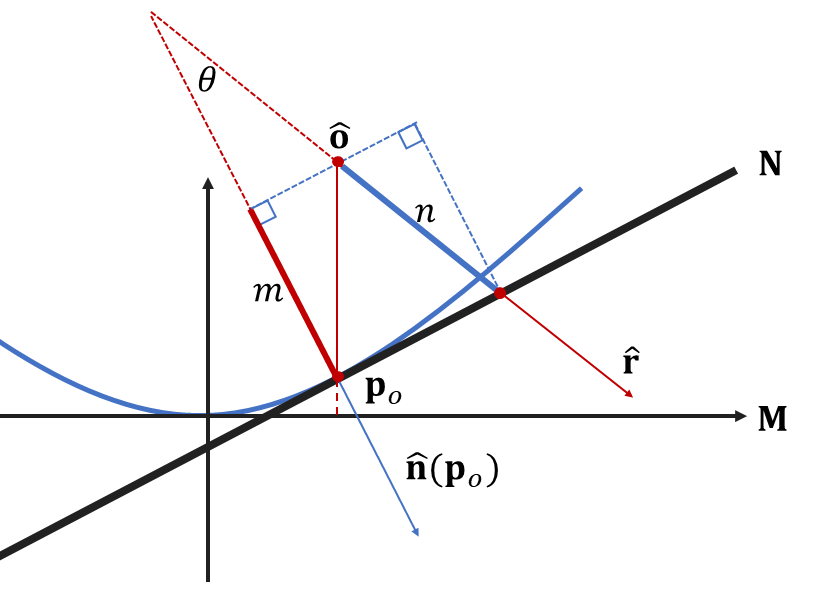}
    \caption{Illustration of approximate intersections. The blue thick line represents the depth of the approximate intersection.}
    \label{fig:numeral_demo}
    \vspace{-15pt}
\end{figure}
\subsection*{Details of Ray-splat Intersection}
\vspace{-3pt}
Given the camera center $\hat{\mathbf{o}} = [\hat{o}^1, \hat{o}^2, \hat{o}^3]^T$ and ray direction $\hat{\mathbf{r}} = [\hat{r}^1, \hat{r}^2, \hat{r}^3]^T$ in the Gaussian coordinate system, the ray can be expressed as:

\begin{equation}
    \hat{\textbf{x}}=\left[
\begin{matrix}
\hat{o}^1+t\hat{r}^1\\
\hat{o}^2+t\hat{r}^2\\
\hat{o}^3+t\hat{r}^3
\end{matrix}
\right]
\label{eq:ray}
\end{equation}

Here, $t$ denotes the depth. 
By solving the ray equation \ref{eq:ray} together with the quadric surface equation 7 in the main text, we find two intersection points:
\begin{equation}
\begin{aligned}
&t^2[\frac{\mathrm{sign}(s_1)}{s_1^2}(\hat{r}^1)^2+\frac{\mathrm{sign}(s_2)}{s_2^2}(\hat{r}^2)^2]+\\&t[\frac{\mathrm{sign}(s_1)}{s_1^2}2\hat{o}^1\hat{r}^1+\frac{\mathrm{sign}(s_2)}{s_2^2}2\hat{o}^2\hat{r}^2-\frac{1}{s_3}\hat{r}^3]+\\&[\frac{\mathrm{sign}(s_1)}{s_1^2}(\hat{o}^1)^2+\frac{\mathrm{sign}(s_2)}{s_2^2}(\hat{o}^2)^2-\frac{1}{s_3}\hat{o}^3]=0\\
&At^2+Bt+C=0\\
&t_{n,f}=\frac{-B\pm\mathrm{sign}(A)\cdot\sqrt{B^2-4AC}}{2A}
\end{aligned}
\label{eq:intersection}
\end{equation}
However, Equation \ref{eq:intersection} encounters numerical issues when $|A|$ is very small, particularly during backpropagation, as a small denominator can cause floating-point overflow, resulting in invalid values such as NaN or Inf.
Upon examination, we observe that when $A$ is especially small, it corresponds to cases where the quadric surface is nearly flat or when the ray is almost perpendicular to the $s_1\times s_2$ plane.
In such cases, we can ignore the quadratic term, reducing Equation \ref{eq:intersection} to $Bt + C = 0$, or $t = -\frac{C}{B}$. 
Geometrically, this solution represents the depth of the intersection between the ray and the tangent plane determined by the point where the perpendicular line from the camera center meets the quadric surface, as indicated by the thick blue line $n$ in Fig \ref{fig:numeral_demo}.

Let $\mathbf{M} = s_1 \times s_2$ be the horizontal plane in the Gaussian local coordinate system. 
Let $\mathbf{p}_o = [\hat{o}_1, \hat{o}_2, s_3(\frac{\hat{o}_1^2}{s_1^2} + \frac{\hat{o}_2^2}{s_2^2})]^T$ represent the intersection of the perpendicular line from the camera center to $\mathbf{M}$ and the quadric surface. 
The normal of the tangent plane $\mathbf{N}$ at the intersection point can be expressed as:
\vspace{-5pt}
\begin{equation*}
    \mathbf{\hat{n}}(\mathbf{p}_o)=[\frac{2\mathrm{sign}(s_1)\hat{o}_1}{s_1^2},\frac{2\mathrm{sign}(s_2)\hat{o}_2}{s_2^2},-\frac{1}{s_3}]^T
\end{equation*}
The projection of $(\mathbf{p}_o - \hat{\mathbf{o}})$ onto the normal direction $\mathbf{n}$ is:
\begin{equation*}
    m=\frac{\hat{\mathbf{n}}(\mathbf{p}_o)\cdot(\mathbf{p}_o-\mathbf{\hat{o}})}{||\hat{\mathbf{n}}(\mathbf{p}_o)||}
\end{equation*}
As indicated by the thick red line segment in the Fig \ref{fig:numeral_demo}. 
On the other hand, the cosine of the angle between the normal vector and the viewing direction can be expressed as:
\begin{equation*}
\mathrm{cos}\theta=\frac{\mathbf{\hat{r}}\cdot\mathbf{\hat{n}}(\mathbf{p}_o)}{||\mathbf{\hat{r}}||\cdot||\mathbf{\hat{n}}(\mathbf{p}_o)||}
\end{equation*}
Noting that $||\hat{\mathbf{r}}||=1$, the length of the line from the camera center along the viewing direction to the intersection with the tangent plane is given by:
\begin{equation*}
    \begin{aligned}
n&=\frac{m}{\cos\theta}=\frac{\hat{\mathbf{n}}(\mathbf{p}_o)\cdot(\mathbf{p}_o-\mathbf{\hat{o}})}{\mathbf{\hat{r}}\cdot\mathbf{\hat{n}}(\mathbf{p}_o)}
=-\frac{C}{B}
\end{aligned}
\end{equation*}
It is evident that when the surface is nearly flat or the viewing direction is almost perpendicular to the horizontal plane, the approximate solution nearly coincides with the exact solution. Therefore, for cases where $|A| < 1\mathrm{e}-6$, we approximate the solution using the method described above.
\vspace{-5pt}
\subsection*{Derivation of Geodesic Arc Length}
\vspace{-3pt}
For the geodesic arc length formula:
\begin{equation}
    \begin{aligned}
l(a,\rho_0)&=\int\sqrt{1+(2at)^2}dt\\
\end{aligned}
\label{eq:geodesic}
\end{equation}
Let $u=2at,x=\arctan(u)$, then Equation \ref{eq:geodesic} becomes:
\begin{equation}
    \begin{aligned}
l(a,\rho_0)&=\int\frac{\sqrt{1+u^2}}{2a}du\\
&=\frac{1}{2a}\int\sqrt{1+\tan^2(x)}d\tan(x)\\
&=\frac{1}{2a}\int\sec^3(x)dx\\
\end{aligned}
\label{eq:part}
\end{equation}
Equation \ref{eq:part} can be solved using integration by parts:
\begin{equation*}
    \begin{aligned}
        \int\sec^3(x)dx
&=\tan(x)\sec(x)-\int\sec^3(x)dx\\&+\int\sec(x)dx\\
    \end{aligned}
\end{equation*}
Using $\int\sec(x)dx=\ln|\sec(x)+\tan(x)|+C$, we have:
\begin{equation*}
    \begin{aligned}
\int\sec^3(x)dx&=\frac{\tan(x)\sec(x)+\ln|\sec(x)+\tan(x)|}{2}\\
&=(u\sqrt{1+u^2}+\ln(u+\sqrt{1+u^2}))/2\\
\Rightarrow l(a,\rho_0)&=\frac{u\sqrt{1+u^2}+\ln(u+\sqrt{1+u^2})}{4a}
\end{aligned}
\end{equation*}
\subsection*{Details of Curvature}
\vspace{-3pt}
Here, we compute the Gaussian curvature analytically using a standard differential geometry approach \cite{stoker2011differential}. 
Given the intersection point $\hat{\mathbf{p}}_0 = [\hat{x}_0, \hat{y}_0, \hat{z}_0]^T$, we simplify Equation 7 as $\hat{z} = \lambda_1 \hat{x}^2 + \lambda_2 \hat{y}^2$. The partial derivatives at $\hat{\mathbf{p}}_0$ are:
\begin{equation*}
\begin{aligned}
    x_u=(1,0,2\lambda_1\hat{x}_0),\;\;
    x_v=(0,1,2\lambda_2\hat{y}_0)
\end{aligned}
\end{equation*}
The first fundamental form is:
\begin{equation*}
\begin{aligned}
    E=<x_u,x_u>&=1+4\lambda_1^2\hat{x}_0^2\\
F=<x_u,x_v>&=4\lambda_1\lambda_2\hat{x}_0\hat{y}_0\\
G=<x_v,x_v>&=1+4\lambda_2^2\hat{y}_0^2
\end{aligned}
\end{equation*}
The second fundamental form is:
\begin{equation*}
\begin{aligned}
    n&=\frac{x_u\times x_v}{||x_u\times x_v||}=\frac{(-2\lambda_1\hat{x}_0,-2\lambda_2\hat{y}_0,1)}{\sqrt{1+4\lambda_1^2\hat{x}_0^2+4\lambda_2^2\hat{y}_0^2}}\\
x_{uu}&=(0,0,2\lambda_1),\;
x_{uv}=(0,0,0),\;
x_{vv}=(0,0,2\lambda_2)\\
L&=<n,x_{uu}>=\frac{2\lambda_1}{\sqrt{1+4\lambda_1^2\hat{x}_0^2+4\lambda_2^2\hat{y}_0^2}}\\
M&=<n,x_{uv}>=0\\
N&=<n,x_{vv}>=\frac{2\lambda_2}{\sqrt{1+4\lambda_1^2\hat{x}_0^2+4\lambda_2^2\hat{y}_0^2}}
\end{aligned}
\end{equation*}
Finally, the Gaussian curvature can be computed as:
\begin{equation*}
    K=\frac{LN-M^2}{EG-F^2}=\frac{\frac{4\lambda_1\lambda_2}{1+4\lambda_1^2\hat{x}_0^2+4\lambda_2^2\hat{y}_0^2}}{1+4\lambda_1^2\hat{x}_0^2+4\lambda_2^2\hat{y}_0^2}
\end{equation*}

\begin{figure*}[!t]
    \centering

    \begin{minipage}{0.05\linewidth}
        \rotatebox{90}{\textbf{scan37}}
    \end{minipage}
    \begin{minipage}{0.170\linewidth}
        \includegraphics[width=\linewidth]{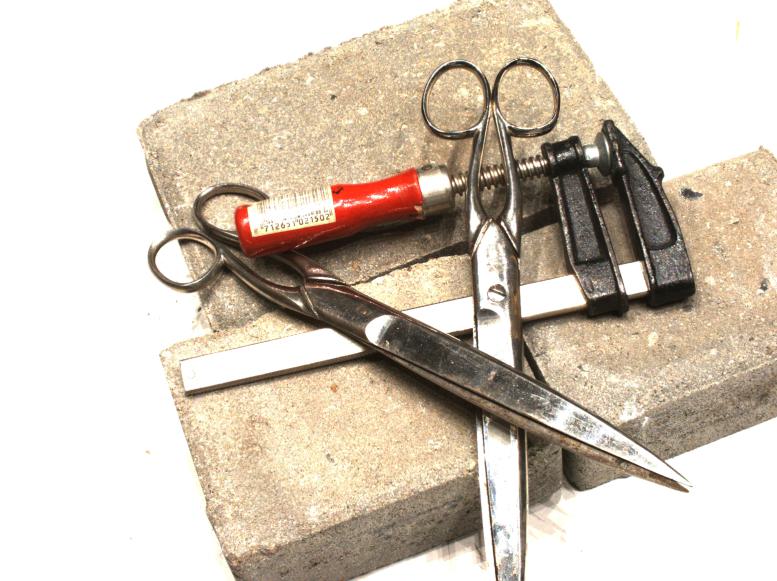}
    \end{minipage}
    \begin{minipage}{0.170\linewidth}
    \includegraphics[width=\linewidth]{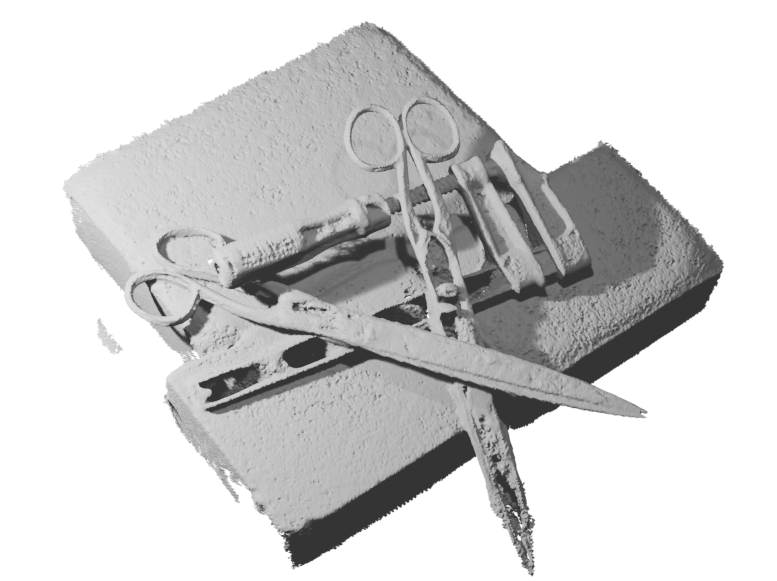}
    \end{minipage}
    \begin{minipage}{0.170\linewidth}
        \includegraphics[width=\linewidth]{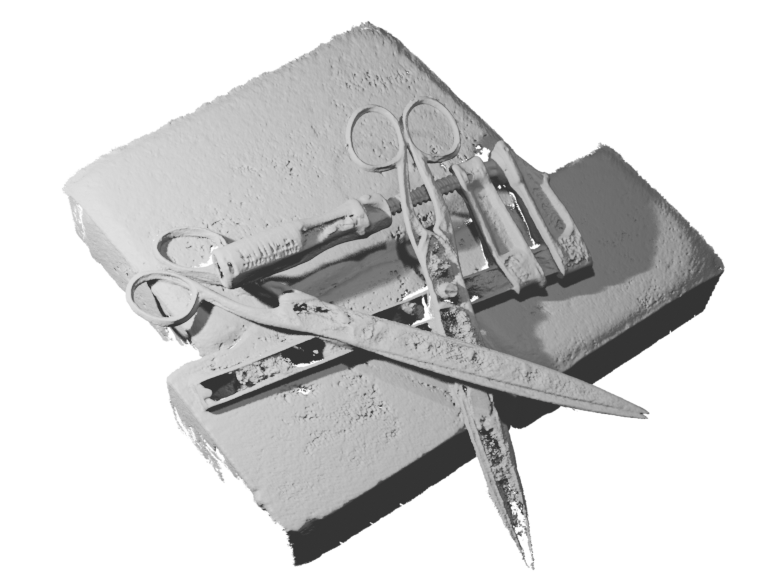}
    \end{minipage}
    \begin{minipage}{0.170\linewidth}
        \includegraphics[width=\linewidth]{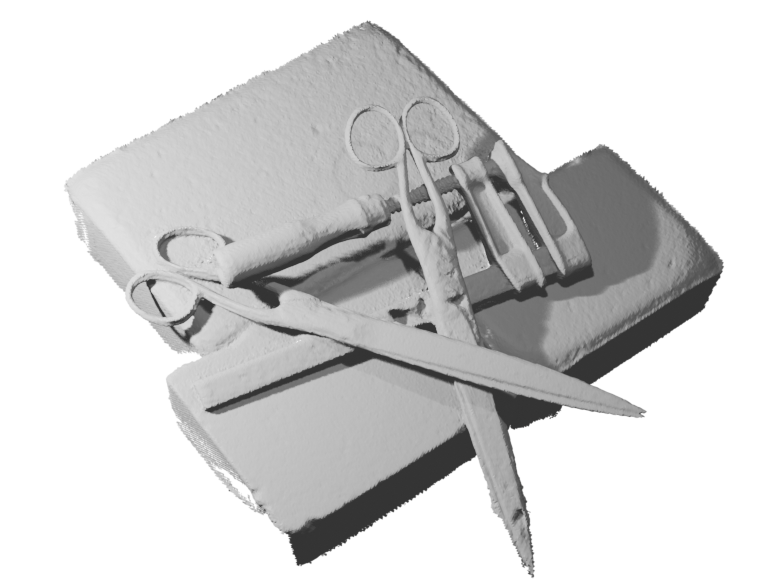}
    \end{minipage}
    \begin{minipage}{0.170\linewidth}
        \includegraphics[width=\linewidth]{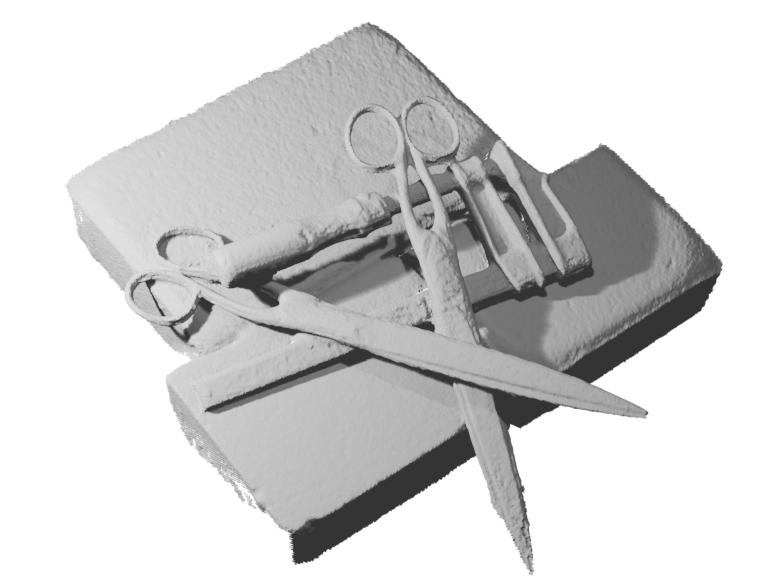}
    \end{minipage}

    \begin{minipage}{0.05\linewidth}
        \rotatebox{90}{\textbf{scan55}}
    \end{minipage}
    \begin{minipage}{0.170\linewidth}
        \includegraphics[width=\linewidth]{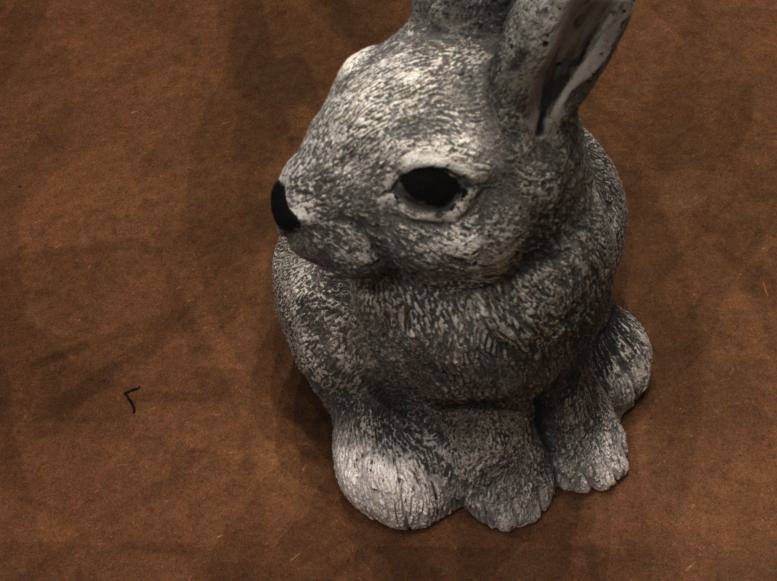}
    \end{minipage}
    \begin{minipage}{0.170\linewidth}
    \includegraphics[width=\linewidth]{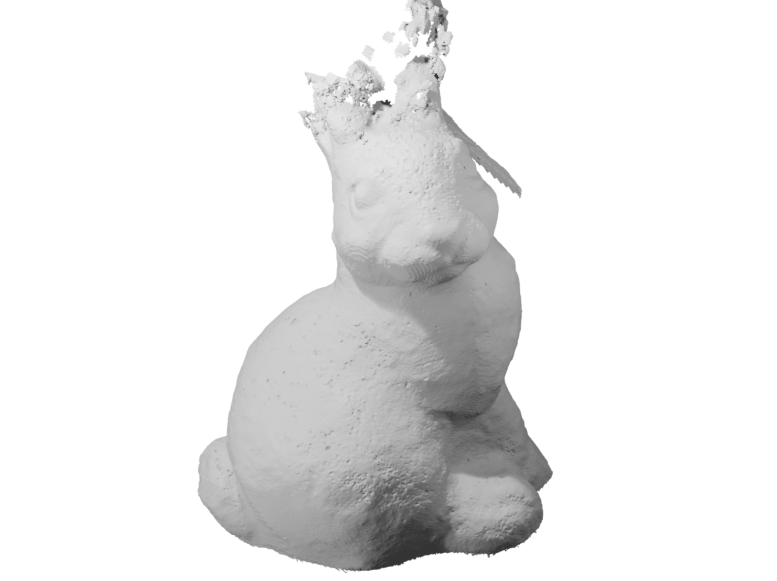}
    \end{minipage}
    \begin{minipage}{0.170\linewidth}
        \includegraphics[width=\linewidth]{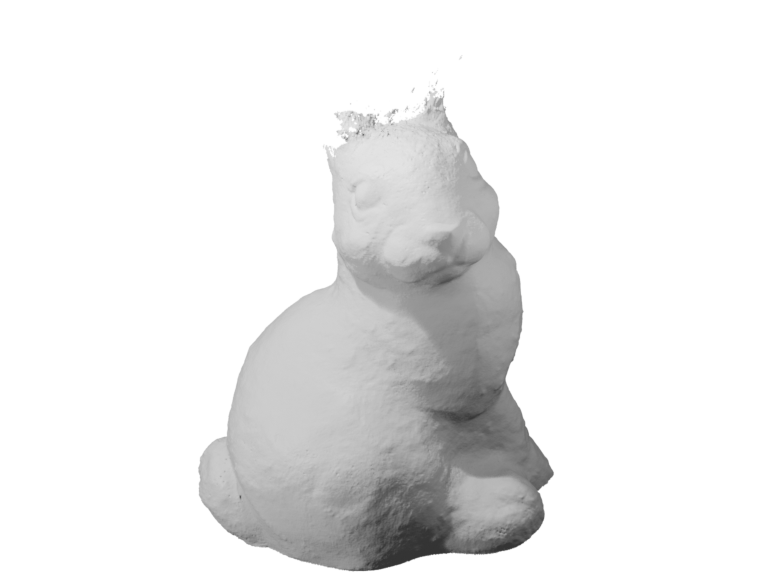}
    \end{minipage}
    \begin{minipage}{0.170\linewidth}
        \includegraphics[width=\linewidth]{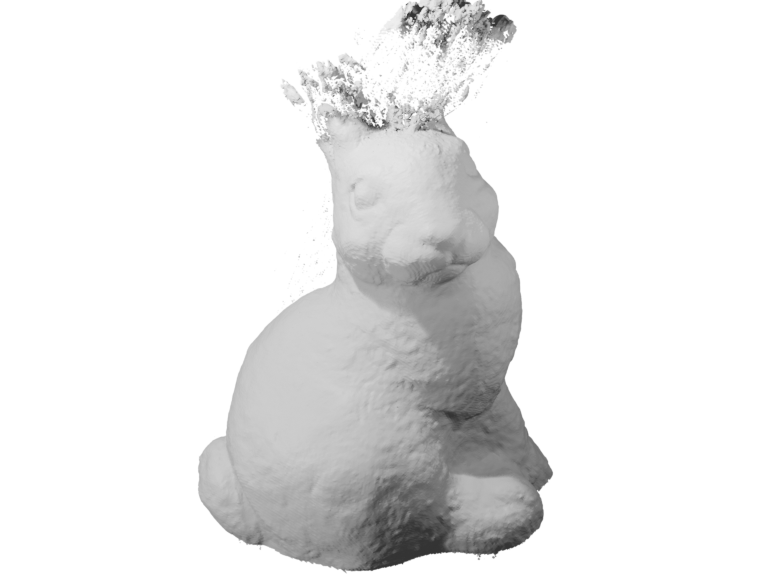}
    \end{minipage}
    \begin{minipage}{0.170\linewidth}
        \includegraphics[width=\linewidth]{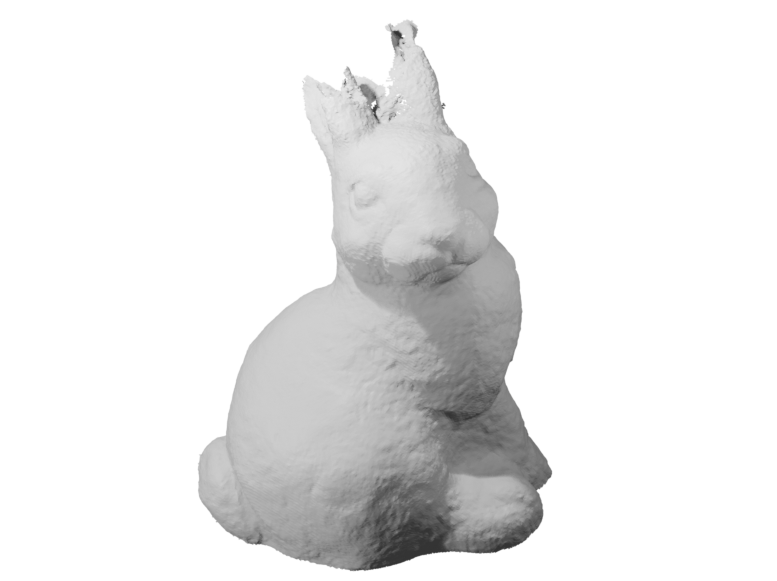}
    \end{minipage}

    \begin{minipage}{0.05\linewidth}
        \rotatebox{90}{\textbf{scan63}}
    \end{minipage}
    \begin{minipage}{0.170\linewidth}
        \includegraphics[width=\linewidth]{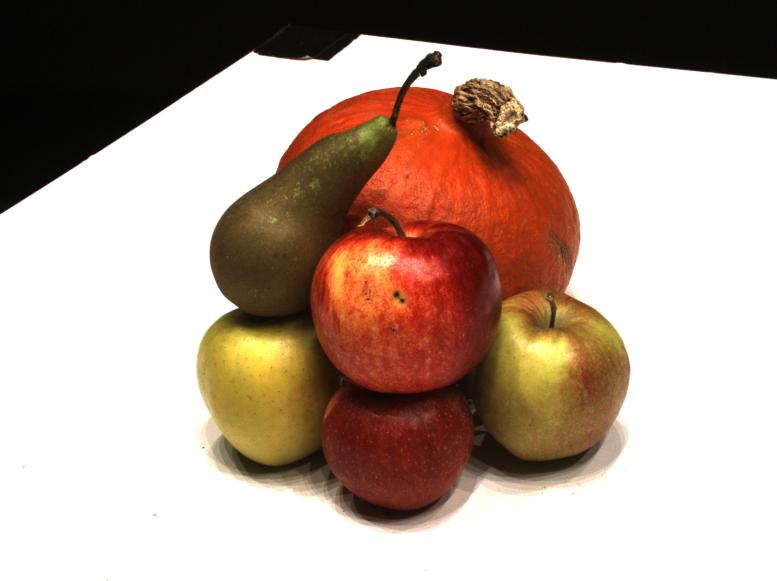}
    \end{minipage}
    \begin{minipage}{0.170\linewidth}
    \includegraphics[width=\linewidth]{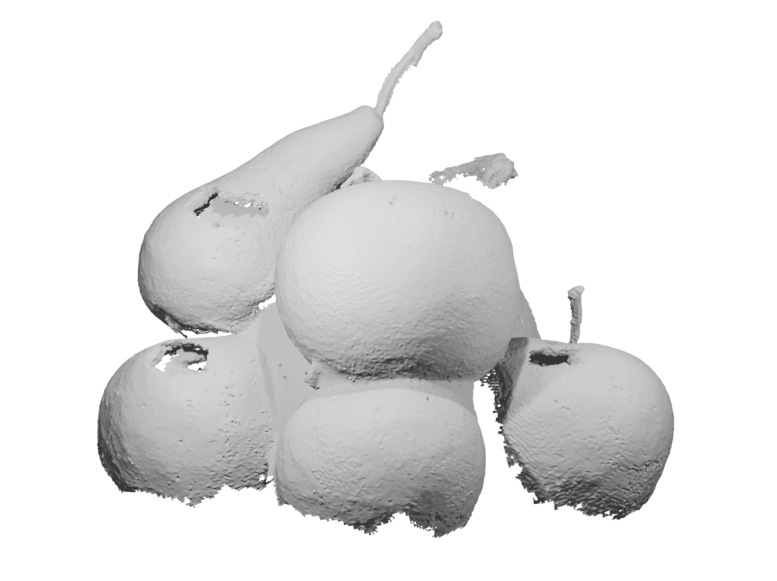}
    \end{minipage}
    \begin{minipage}{0.170\linewidth}
        \includegraphics[width=\linewidth]{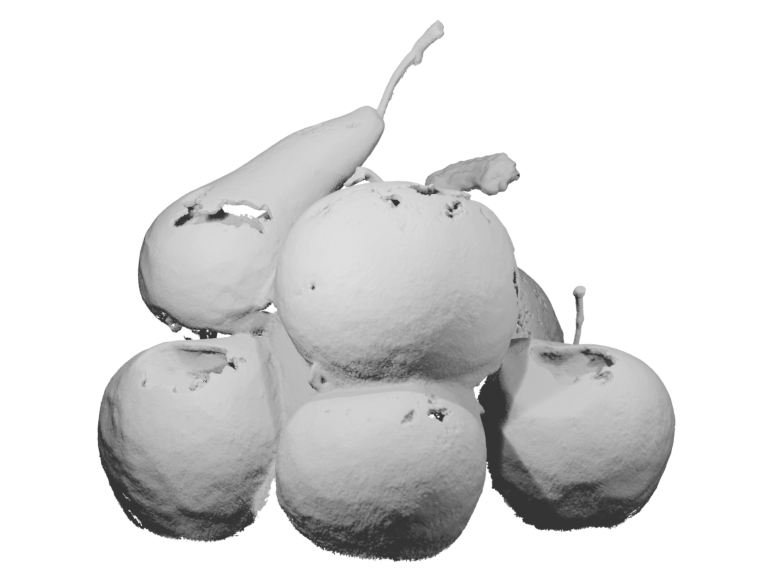}
    \end{minipage}
    \begin{minipage}{0.170\linewidth}
        \includegraphics[width=\linewidth]{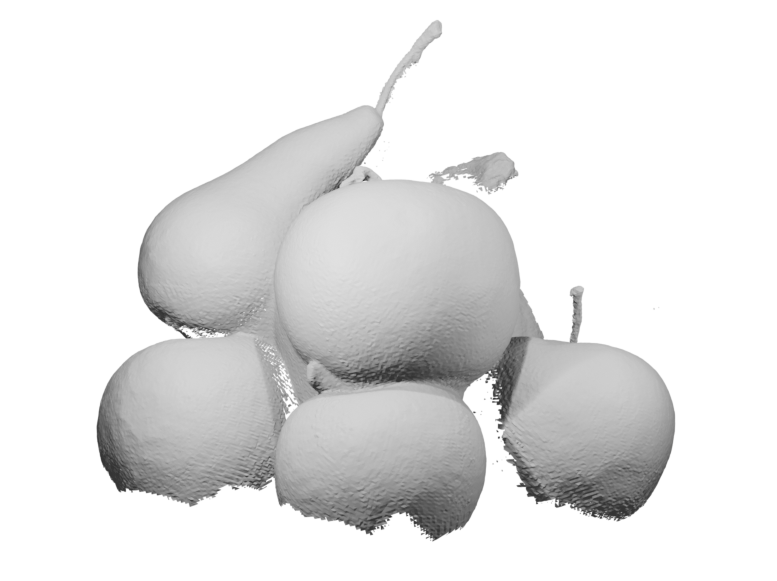}
    \end{minipage}
    \begin{minipage}{0.170\linewidth}
        \includegraphics[width=\linewidth]{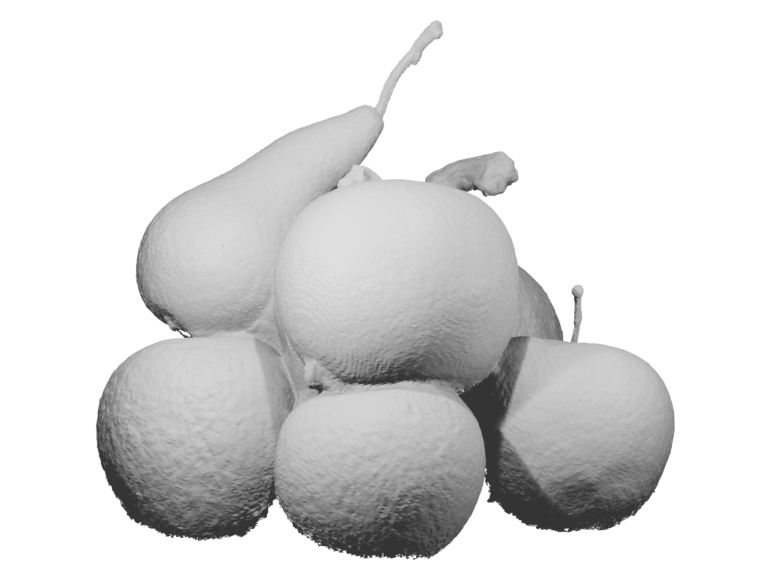}
    \end{minipage}

    \begin{minipage}{0.05\linewidth}
        \rotatebox{90}{\textbf{scan69}}
    \end{minipage}
    \begin{minipage}{0.170\linewidth}
        \includegraphics[width=\linewidth]{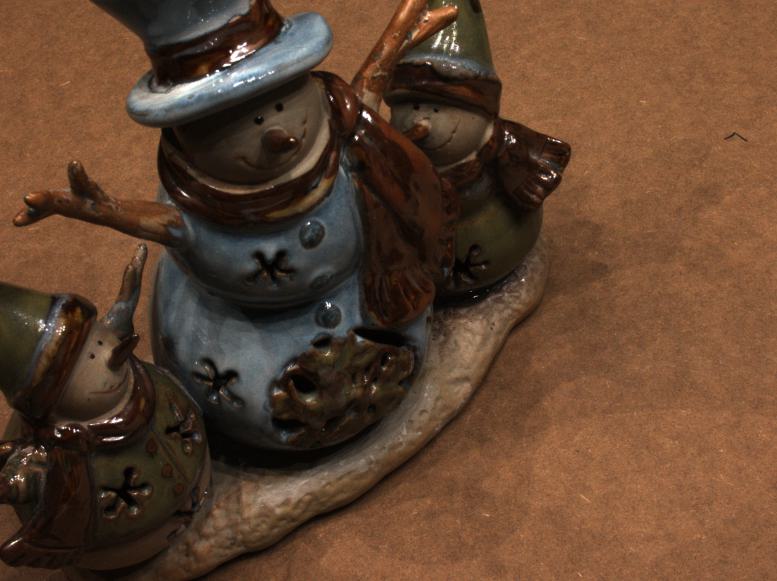}
    \end{minipage}
    \begin{minipage}{0.170\linewidth}
    \includegraphics[width=\linewidth]{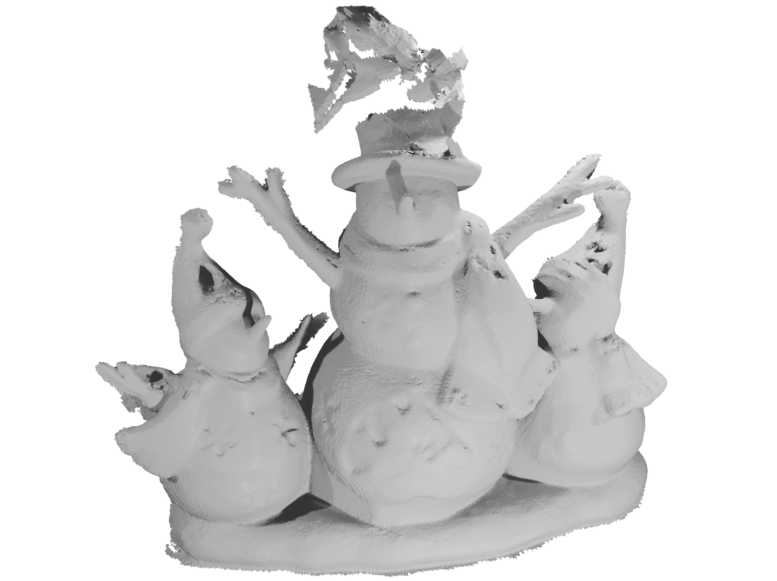}
    \end{minipage}
    \begin{minipage}{0.170\linewidth}
        \includegraphics[width=\linewidth]{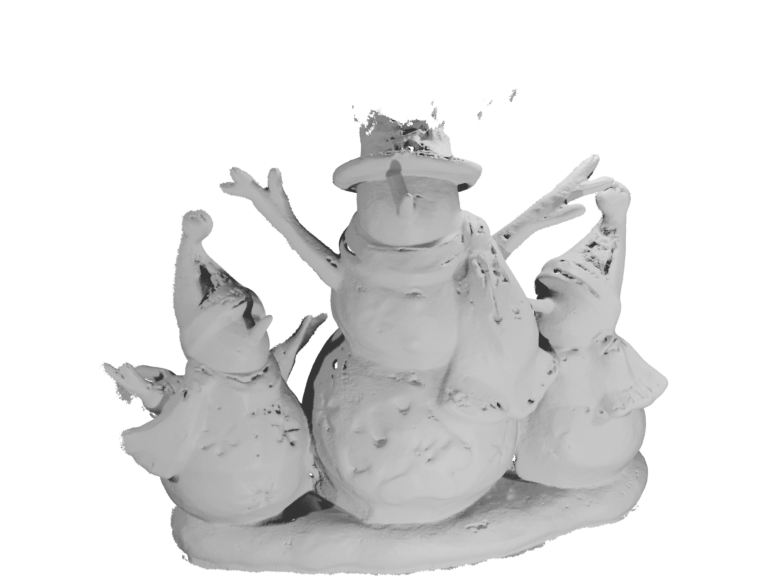}
    \end{minipage}
    \begin{minipage}{0.170\linewidth}
        \includegraphics[width=\linewidth]{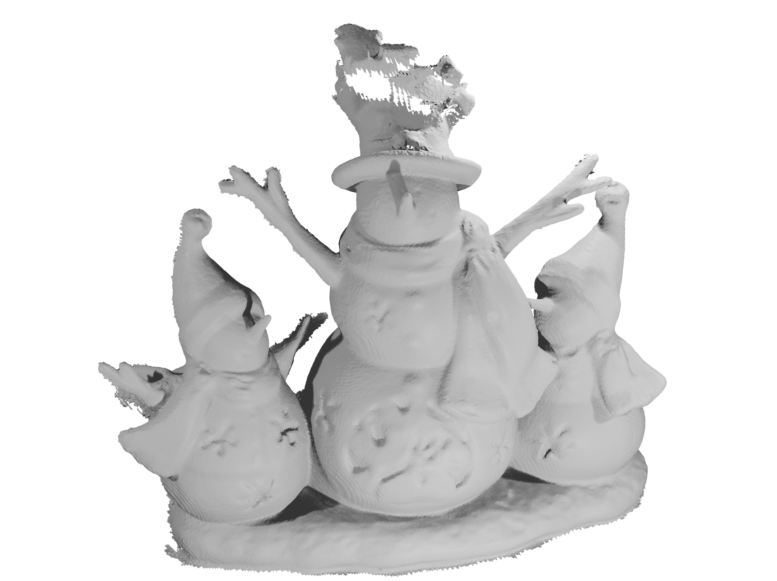}
    \end{minipage}
    \begin{minipage}{0.170\linewidth}
        \includegraphics[width=\linewidth]{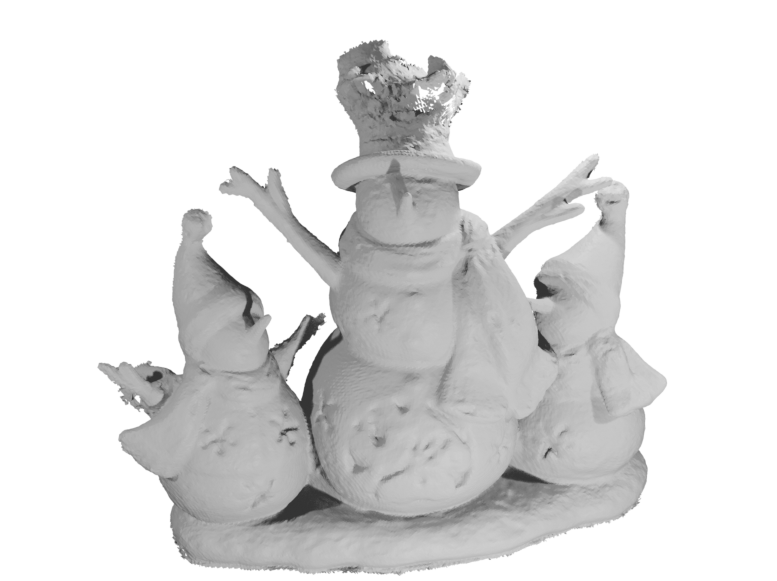}
    \end{minipage}

    \begin{minipage}{0.05\linewidth}
        \rotatebox{90}{\textbf{scan110}}
    \end{minipage}
    \begin{minipage}{0.170\linewidth}
        \includegraphics[width=\linewidth]{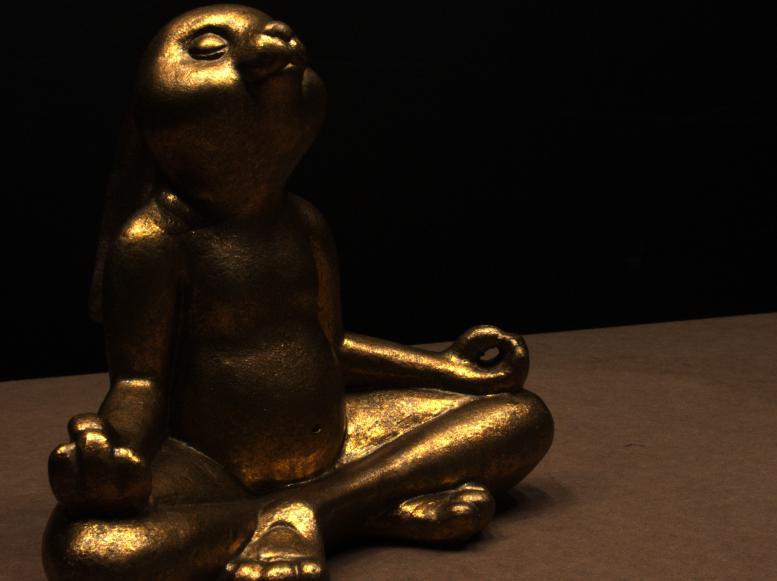}
        \caption*{Reference Image}
    \end{minipage}
    \begin{minipage}{0.170\linewidth}
    \includegraphics[width=\linewidth]{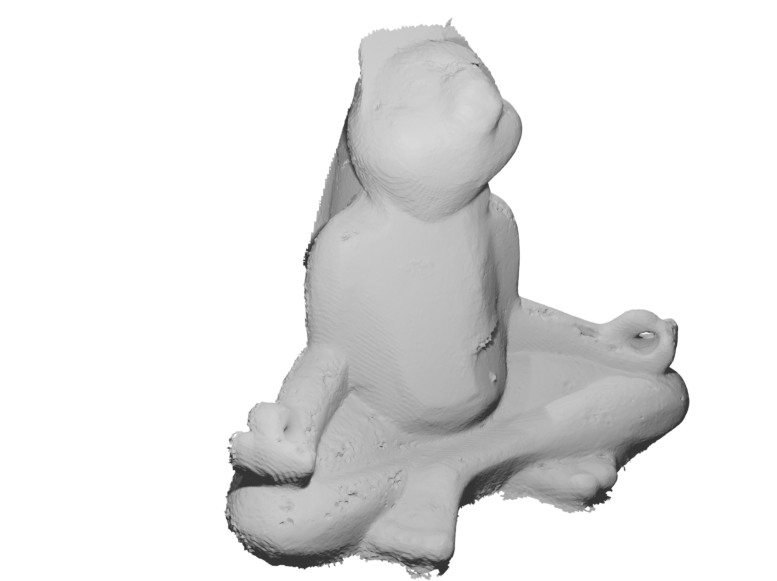}
        \caption*{2DGS~\cite{huang20242d}}
    \end{minipage}
    \begin{minipage}{0.170\linewidth}
        \includegraphics[width=\linewidth]{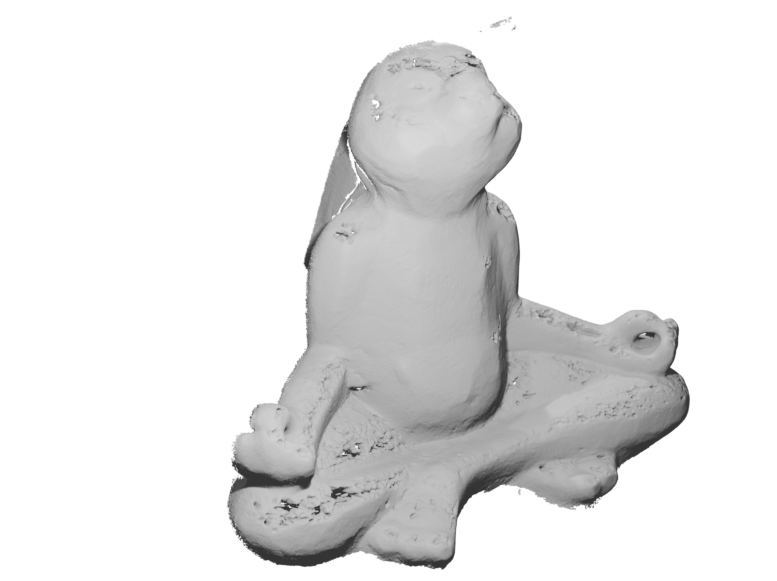}
        
        \caption*{GOF~\cite{yu2024gaussian}}
    \end{minipage}
    \begin{minipage}{0.170\linewidth}
        \includegraphics[width=\linewidth]{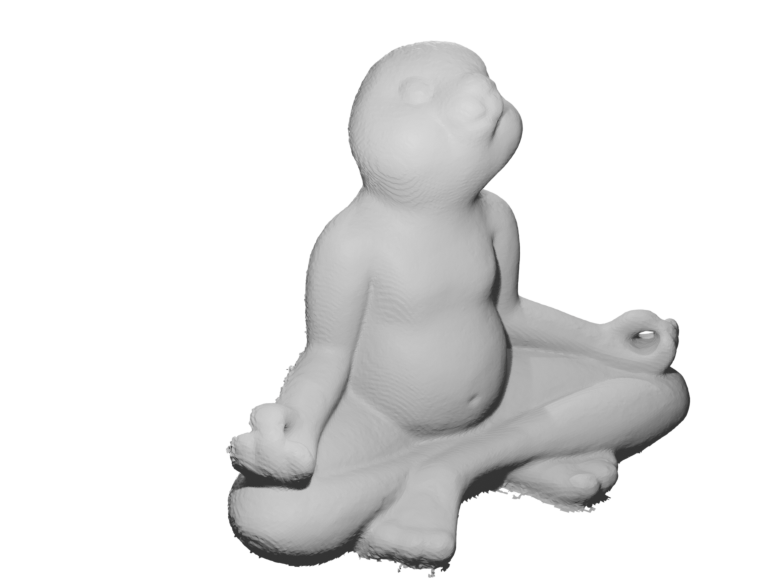}
        \caption*{PGSR~\cite{chen2024pgsr}}
    \end{minipage}
    \begin{minipage}{0.170\linewidth}
        \includegraphics[width=\linewidth]{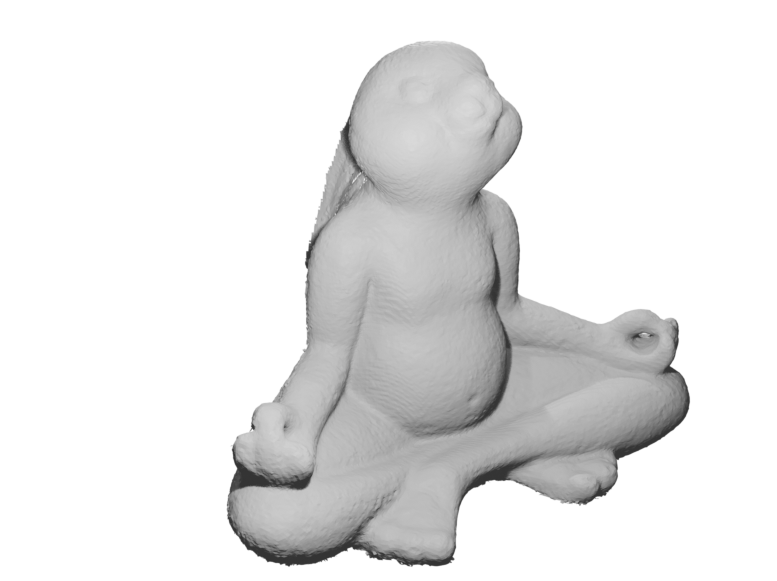}
        \caption*{Ours}
    \end{minipage}

    \caption{Qualitative geometric reconstruction comparisons on the DTU dataset~\cite{jensen2014large}. Our method achieves reconstructions of higher quality and greater detail. }
    \label{fig:supp:qualitative_dtu}
\end{figure*}

\begin{figure*}[!t]
    \centering
    \includegraphics[width=0.95\linewidth]{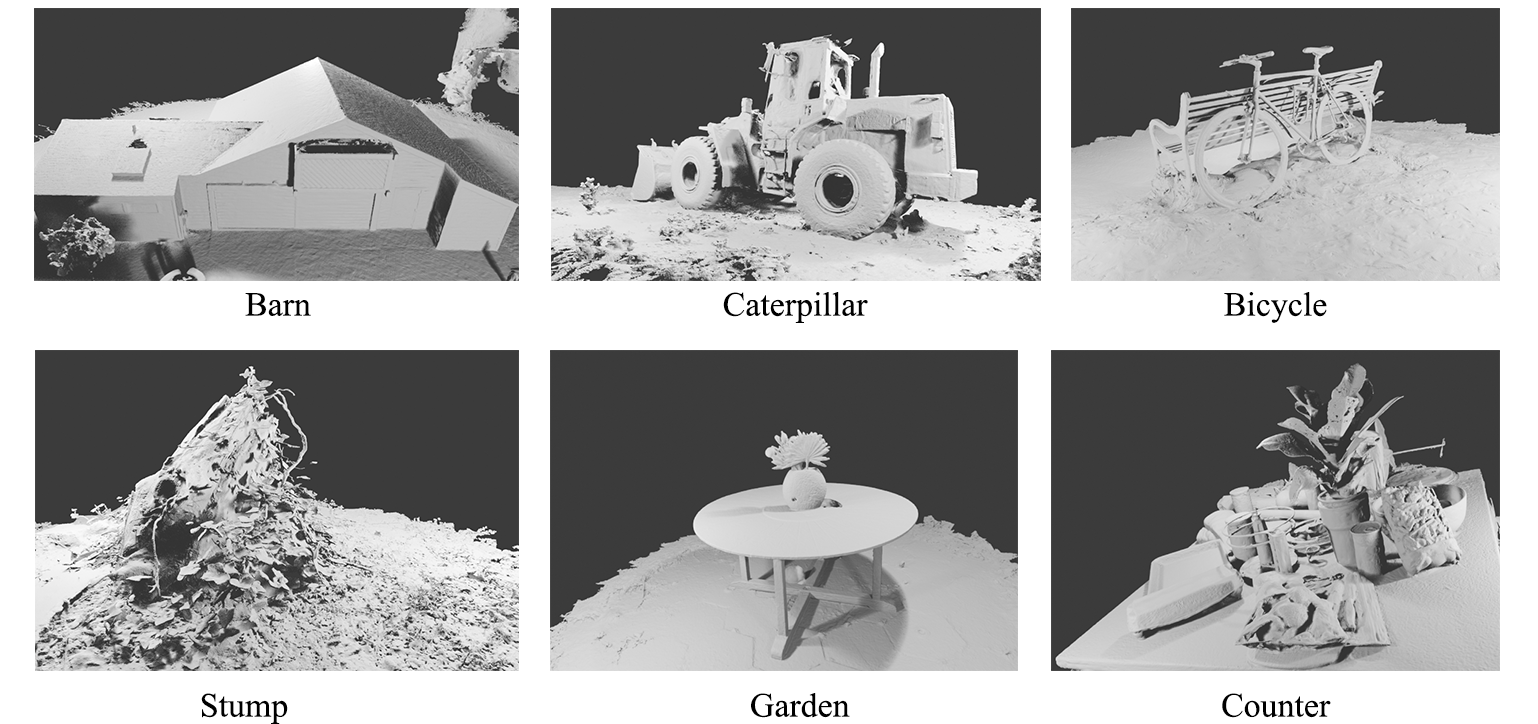}
    \caption{Qualitative geometry results for the Tanks and Temples dataset \cite{knapitsch2017tanks} and Mip-NeRF 360 dataset \cite{barron2022mip}.}
    \label{fig:supp_TNT}
\end{figure*}

\begin{figure*}[!t]
    \centering
    \includegraphics[width=0.8\linewidth]{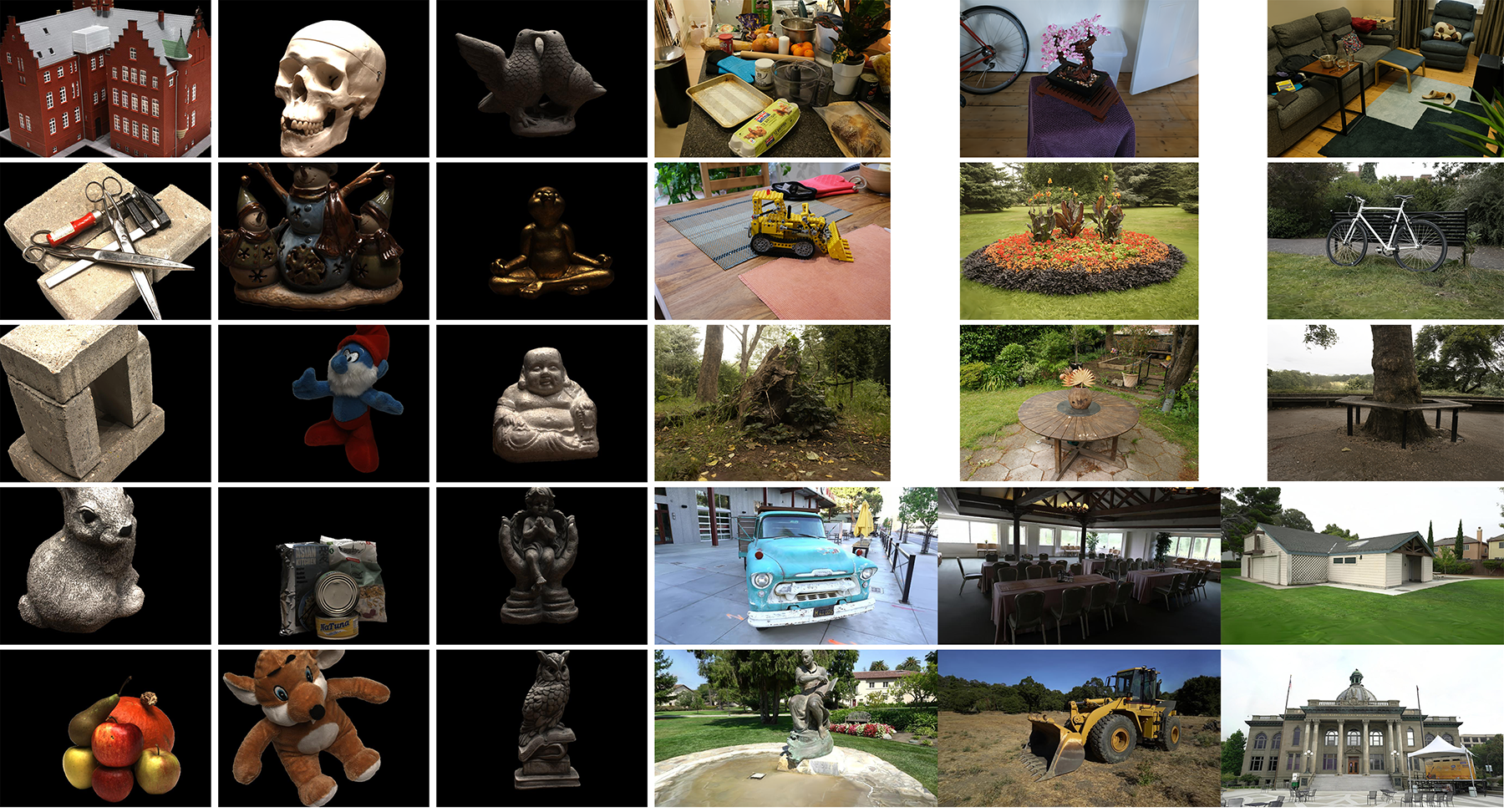}
    \caption{Qualitative appearance results for the DTU, TNT and Mip-NeRF 360 dataset.}
    \label{fig:supp:appearance}
\end{figure*}

\begin{figure*}[!t]
    \centering
    \includegraphics[width=0.8\linewidth]{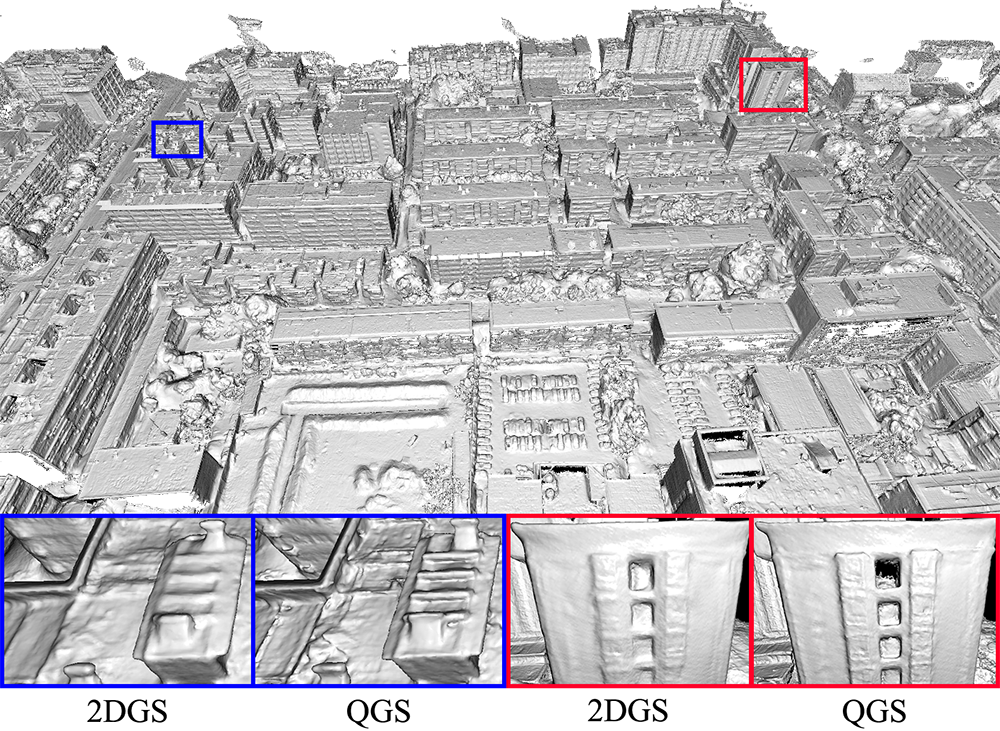}
    \caption{Qualitative geometry result for the urban scene captured from aerial view, involving over 1,000 images. In the subfigure, the left shows the results of 2DGS~\cite{huang20242d}, while the right shows the results of QGS.}
    \label{fig:supp_large}
\end{figure*}

\noindent\textbf{Addtional Results. }
In this section, we present additional qualitative results of QGS in both indoor and outdoor scenarios. 
Figure \ref{fig:supp:qualitative_dtu} presents a comparison of QGS with 2DGS~\cite{huang20242d}, GOF~\cite{yu2024gaussian}, and PGSR~\cite{chen2024pgsr} on the DTU dataset, illustrating that QGS captures more geometric details. 
QGS also achieves accurate reconstructions in indoor and outdoor scenes, as shown in Fig \ref{fig:supp_TNT} with results from the TNT and Mip-NeRF 360 datasets. 
Rendering results for all three datasets are shown in Fig \ref{fig:supp:appearance}. 
Finally, we qualitatively tested our method on an urban scene captured from aerial view. 
As shown in Fig \ref{fig:supp_large}, QGS captures more building details than 2DGS \cite{huang20242d} in aerial views. 

